\documentclass[journal]{IEEEtran}

\usepackage{cite}
\usepackage{amsmath,amssymb,amsfonts}
\usepackage{graphicx}
\usepackage{booktabs}
\usepackage{multirow}
\usepackage{array}
\usepackage{siunitx}
\usepackage{subfigure} 
\usepackage{hyperref}

\title{Domain Adaptation for Big Data in Agricultural Image Analysis: A Comprehensive Review}

\author{
Xing~Hu\thanks{Xing Hu is with the School of Optical-Electrical and Computer Engineering, University of Shanghai for Science and Technology, Shanghai 200093, China (e-mail: huxing@usst.edu.cn).}, 
Siyuan~Chen\thanks{Siyuan Chen is with the School of Optical-Electrical and Computer Engineering, University of Shanghai for Science and Technology, Shanghai 200093, China (e-mail: 242250440@st.usst.edu.cn).}, 
Qianqian~Duan\thanks{Qianqian Duan is with the School of Electronics and Electrical Engineering, Shanghai University of Engineering Science, Shanghai 201620, China (e-mail: dqq1019@163.com).}, 
Choon Ki~Ahn~\IEEEmembership{Senior Member,~IEEE}\thanks{Choon Ki AhnSchool of Electrical Engineering, Korea University, 145 Anam-ro, Seongbuk-gu, Seoul, 02841, South Korea (e-mail: ckahn@korea.ac.kr).},
Huiliang~Shang\thanks{Huiliang Shang is affiliated with the Future Information Innovation Institute, Fudan University, Shanghai, China (e-mail: shanghl@fudan.edu.cn) .}, 
and~Dawei~Zhang~\IEEEmembership{Senior Member,~IEEE}\thanks{Dawei Zhang is with the School of Optical-Electrical and Computer Engineering, University of Shanghai for Science and Technology, Shanghai 200093, China (e-mail: dwzhang@usst.edu.cn) .}
}

\begin{document}
 \maketitle

\begin{abstract}
With the wide application of computer vision in agriculture, image analysis has become the key to tasks such as crop health monitoring and pest detection. However, the significant domain shifts caused by environmental changes, different crop types, and diverse data acquisition methods seriously hinder the generalization ability of the model in cross-region, cross-season, and complex agricultural scenarios. This paper explores how domain adaptation (DA) techniques can address these challenges to improve cross-domain transferability in agricultural image analysis. DA is considered a promising solution in the case of limited labeled data, insufficient model adaptability, and dynamic changes in the field environment. This paper systematically reviews the latest advances in DA in agricultural images in recent years, focusing on application scenarios such as crop health monitoring, pest and disease detection, and fruit identification, in which DA methods have significantly improved cross-domain performance. We categorize DA methods into shallow learning and deep learning methods, including supervised, semi-supervised and unsupervised strategies, and pay special attention to the adversarial learning-based techniques that perform well in complex scenarios. In addition, this paper also reviews the main public datasets of agricultural images, and evaluates their advantages and limitations in DA research. Overall, this study provides a complete framework and some key insights that can be used as a reference for the research and development of domain adaptation methods in future agricultural vision tasks.
\end{abstract}

\section*{Keywords}
Domain Adaptation, Deep Learning, Agricultural Image Analysis, Cross-Domain Transfer, Adversarial Learning


\section{Introduction}
In agricultural image analysis, the high heterogeneity of data acquisition environment and the extreme scarcity of labeled data severely limit the generalization ability of the model in different application scenarios. Among these challenges, the \textbf{domain shift problem} - that is, the distribution difference between the source domain and the target domain data - has become a key obstacle to the deployment of intelligent agricultural vision systems~\cite{123,126}. In order to solve this problem, \textbf{domain adaptation (DA)} technology has attracted wide attention in agricultural image analysis in recent years, and its goal is to improve the robustness and transferability of the model on data sources across regions, across time and across devices.

Agricultural image data is usually acquired thru multiple sensors (such as RGB cameras, near-infrared imagers, and hyperspectral sensors), and is highly susceptible to environmental factors such as changes in illumination, seasonal cycles, crop varieties, and soil background. These differences lead to significant distribution shifts between different datasets for the same task, resulting in a severe performance degradation when the models trained under the traditional paradigm are applied in new environments~\cite{141}. For example, models trained in the laboratory or demonstration farms often show a significant decrease in accuracy in real fields, other geographical areas, or different crop growth periods. The application of domain adaptation has been extended beyond crops, such as the work of Dubourvieux et al. on cattle re-identification~\cite{147}. Their multi-domain accumulation strategy addressed issues such as changes in cowshed conditions or camera viewing angles, demonstrating the applicability of DA in livestock management.

In addition, the annotation of agricultural images usually requires the expertise of agronomists and plant protection experts, so the cost is high and the data scalability is limited. Compared with general computer vision tasks, the availability of high-quality labeled datasets in the agricultural field is still extremely limited~\cite{140}. Therefore, directly transferring the pre-trained model in the field of natural images often has poor results, highlighting the necessity of DA technology in reducing label dependence and improving cross-domain adaptability. The core goal of DA is to minimize the difference in feature distribution between the source domain and the target domain, so that the model can have comparable performance in the target domain.

\begin{figure*}[ht]
    \centering
    \includegraphics[width=1\linewidth]{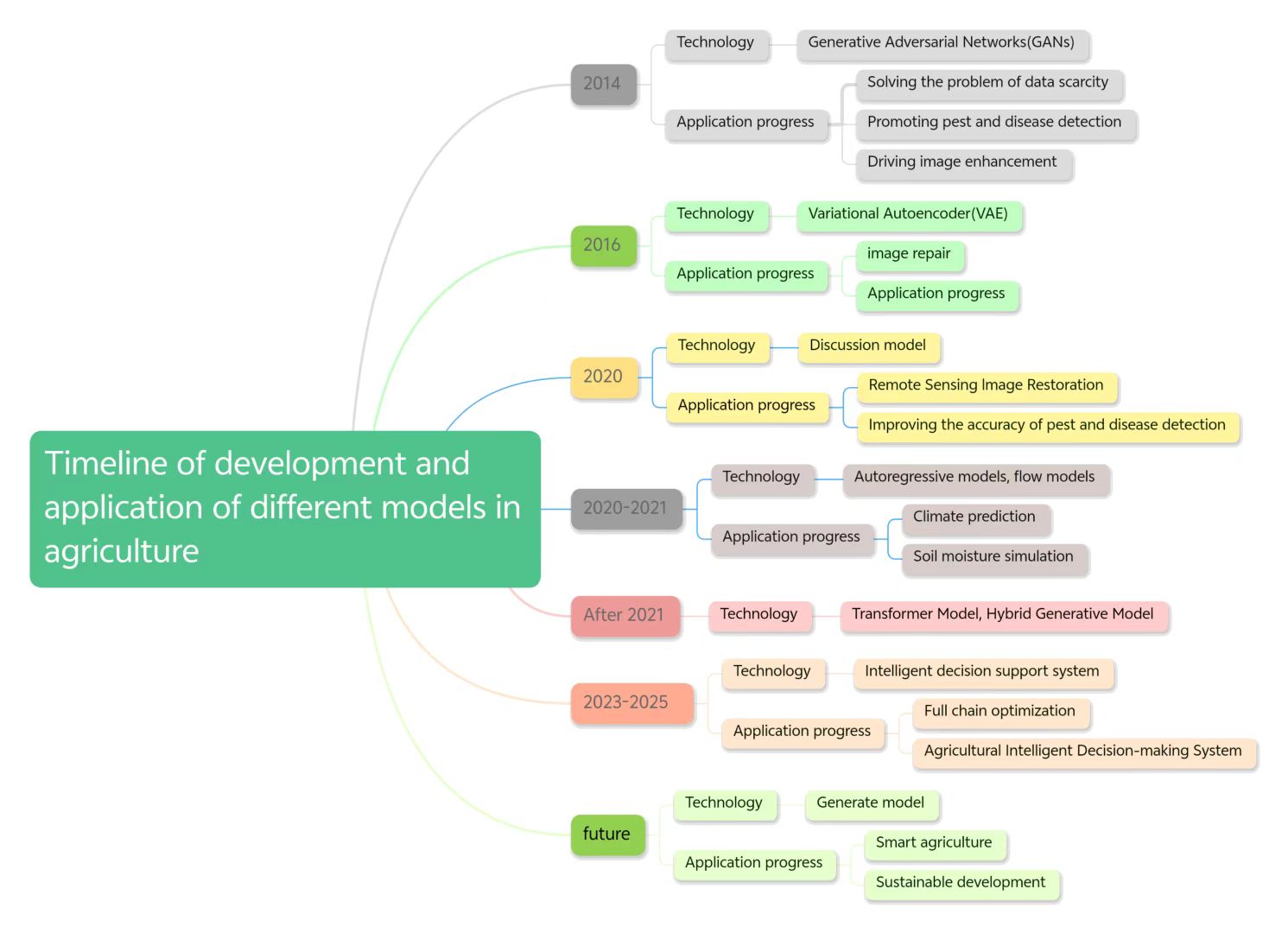}
    \caption{Application of diferant models in agriculture}
    \label{fig20}
\end{figure*}

In recent years, DA has shown significant potential in key applications such as crop pest and disease detection, hyperspectral image analysis, crop growth monitoring, and precision agricultural management. In particular, DA's ability to model distribution shifts under unsupervised or weakly supervised conditions makes it an important means to promote the scale of agricultural visual intelligence. Although there are a large number of reviews on transfer learning and domain adaptation in the field of natural image analysis~\cite{127,128,129,130,131,132,133,134,135,136,137,138}, there are still limited systematic reviews on agricultural image analysis.

To fill this gap, this paper systematically summarizes the current status of domain adaptation methods in agricultural image analysis. We divide the available methods into two groups: shallow learning and deep learning, and further analyze their applications across supervised, semi-supervised, and unsupervised strategies. Special emphasis is placed on the effectiveness of adversarial learning-based adaptation methods in addressing key challenges such as seasonal changes, multi-source data fusion, and scarce labeling. This paper finally provides a methodological reference for the development of the next generation of intelligent agricultural systems.


\section{Background}

\subsection{Domain Shift in Agricultural Image Analysis}

In agricultural image analysis, domain shift is defined as the discrepancy in data distribution between the training set (source domain) and the testing set (target domain).
This distribution difference often leads to good performance of the model in the source domain, but poor performance on new data, which limits its application in practical scenarios. The acquisition conditions of agricultural images are diverse, and are greatly affected by the differences in environment, crop types and sensing equipment. Data from different regions or environments, such as changes in illumination, climate, seasonal cycles, crop varieties, and sensor perspectives, can exacerbate domain shift and reduce cross-domain migration capabilities.

\begin{figure}[ht]
    \centering
    \includegraphics[width=1\linewidth]{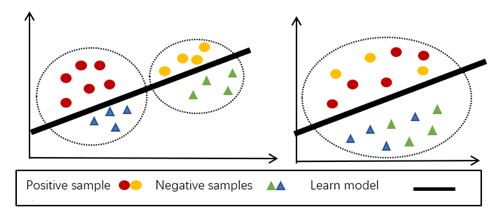}
    \caption{Illustration of source and target data: (left) misalignment of feature distributions is caused by domain shift; (right) improved alignment after applying domain adaptation.}
    \label{fig1}
\end{figure}

As shown in Figure~\ref{fig1}, recent studies have applied \textbf{domain adaptation (DA)} to agricultural vision tasks to address this issue. The goal of DA is to reduce the distribution gap by aligning the feature representations of the source and target domains, thereby achieving more stable performance on heterogeneous datasets.

\begin{figure}[ht]
    \centering
    \includegraphics[width=1\linewidth]{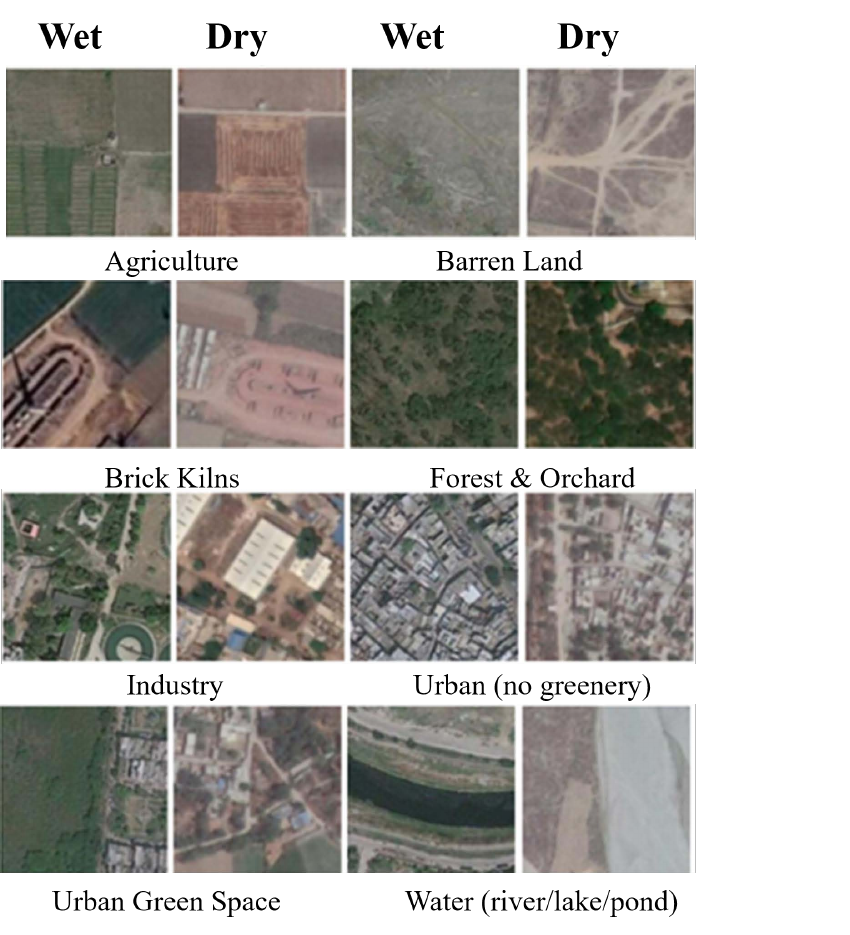}
    \caption{Aerial scenes from Ghaziabad, India, captured during rainy and dry seasons, illustrating domain shift caused by seasonal variation.}
    \label{fig2}
\end{figure}

In practical applications, such as crop classification, images from different regions may show significant differences. Figure~\ref{fig2} shows the aerial images of Ghaziabad, India, collected in the rainy season and the dry season. Seasonal changes in light, vegetation growth, and atmospheric conditions can cause significant distributional differences. These differences can affect the accuracy of agricultural vision tasks such as crop monitoring, pest and disease detection, and yield estimation. This challenge is more prominent in cross-region, cross-season, and cross-sensor applications.

The rapid growth of large-scale multi-source agricultural image data has made domain shift a problem that needs to be solved urgently. In the era of big data, the diversity of data volume and sources makes it more difficult to maintain the stability of model performance. Domain adaptation provides a feasible way to reduce these gaps and help the model obtain more accurate and reliable results in large-scale agricultural applications.

\subsection{Domain Adaptation and Transfer Learning}

Under the framework of transfer learning (TL), the core contains two elements: \textit{domain} and \textit{task}.
A domain is determined by a feature space and its marginal distribution, while a task is composed of a label space and the corresponding prediction function.
The central aim of transfer learning is to reuse knowledge from the source task $T_A$ (domain $A$) and adapt it effectively to the target task $T_B$ (domain $B$).
In different scenarios, the source and target domains may differ in domain, task, or both.

\textit{Domain Adaptation} (DA) is a special subfield of TL, where the feature space and task of the source domain and the target domain are the same, but the marginal distribution is different, i.e., $P_S(X) \neq P_T(X)$.Formally, the joint space of features and labels is denoted as $X \times Y$. The source domain $S$ and the target domain $T$ share the space, but follow different distributions $P_S$ and $P_T$ respectively. Suppose there are $n_S$ labeled samples in the source domain, $D_S = \{ (x_i^S, y_i^S) \}_{i=1}^{n_S}$; there are $n_T$ samples in the target domain, $D_T = \{ x_j^T \}_{j=1}^{n_T}$, which may be partially labeled or completely unlabeled.

The objective of domain adaptation (DA) is to leverage labeled source-domain data to enhance task performance in the target domain when distributional differences exist. The usual implementation method is to align the distribution of the source domain and the target domain, which can be an explicit statistical method or an implicit completion thru the learned feature representation. In large-scale agricultural image analysis and other big data applications, DA plays an important role in dealing with heterogeneous datasets from multiple regions, multiple seasons, and multiple sensors. These datasets usually contain a large number of samples with diverse features, and the model is difficult to maintain stable performance without adaptation. By reducing the distribution gap, the DA method can enhance the cross-domain generalization ability, thereby improving the reliability of data-driven systems in real large-scale applications.

\subsection{Domain Adaptation Problem Setup}

In agriculture, domain adaptation (DA) methods need to deal with large and diverse datasets. The source and target domains often differ due to spatial differences, temporal changes, and multiple sensing modes. Agricultural imagery can come from satellites, unmanned aerial vehicles (UAVs), and ground-based sensors. Differences in climate, crop types, and sensor configurations across regions often result in significant distribution gaps. Bridging these gaps is key to building models that can generalize across environments and remain scalable.

Domain adaptation (DA) methods are commonly categorized based on the quantity of labeled data available in the target domain, namely into supervised, semi-supervised, and unsupervised approaches. Among them, unsupervised domain adaptation (UDA) is particularly important in agriculture. The manual annotation process is expensive and time-consuming, while the datasets often contain millions of unlabeled samples from different regions and seasons.

Early DA research mainly focused on explicitly aligning the source and target domain distributions thru three types of strategies: instance-based alignment (such as sample re-weighting), feature space transformation (such as subspace learning), and classifier parameter adaptation. These methods are collectively referred to as \textbf{Shallow DA}, which usually relies on manual features and statistical alignment. Instance-based methods adjust the contribution of source samples to better match the target distribution; feature-based methods learn a shared subspace thru linear projection; classifier-based methods improve target domain performance thru parameter adjustment. Shallow DA has the characteristics of low computational cost, high efficiency and good interpretability, and is especially valuable in resource-constrained agricultural scenarios.

The capability of domain adaptation (DA) has been significantly enhanced by the progress of deep learning. In 2014, Girshick et al.~\cite{1} proposed the Region-based Convolutional Neural Network (R-CNN) for object detection. Subsequently, Fast R-CNN~\cite{2} and Faster R-CNN~\cite{3} further integrated region proposal, feature extraction and classification into an end-to-end process. In the application of agriculture, Fuentes et al. \cite{4} combined Faster R-CNN with VGGNet/ResNet backbone network to realize the detection and location of tomato diseases and insect pests, and achieved an average accuracy of 85.98
 These case studies highlight the importance of \textbf{deep DA} in large-scale agricultural vision tasks, where feature learning relies on large datasets.

In general, DA methods in agriculture can be roughly divided into two categories: shallow and deep adaptation. Shallow DA mainly reduces the distribution difference thru sample re-weighting, feature alignment and classifier adjustment; while deep DA relies on deep neural networks to learn domain-invariant representations from large-scale data. Depending on the availability of labeled data, both types of methods can be run in supervised, semi-supervised, or unsupervised mode, as summarized in Table~\ref{I}. The two paradigms have their own advantages in terms of computational efficiency, big data scalability, and cross-domain generalization ability, and can form a complementary relationship.

\begin{table*}[ht]
\centering
\caption{Taxonomy of Domain Adaptation Methods in Agricultural Image Analysis}
\label{I}
\renewcommand{\arraystretch}{1.3} 
\setlength{\tabcolsep}{6pt} 
\resizebox{\textwidth}{!}{%
\begin{tabular}{p{3.2cm} p{4cm} p{8.8cm}}
\toprule
\textbf{Category} & \textbf{Subcategory} & \textbf{Representative Methods} \\
\midrule
\multirow{3}{*}{\textbf{Shallow DA}} 
& Instance-based 
& DIAFAN-TL, RHM, HSI, CSSPL \\ \cmidrule(l){2-3}
& Feature-based 
& SA, TSSA, ECMDCM, CropSTGAN, MultiCropGAN, Spatial-Invariant Features, RKHS Features, Sensor-driven Hierarchical DA, PSO-TrAdaBoost, ESMSL, HSL-GM \\ \cmidrule(l){2-3}
& Classifier-based 
& MLCA, BCC, SSGF, MDAF, WRF, CDELM, BHC, DASVM, SD-AL, AMKFL, VSV, EasyTL \\
\midrule
\multirow{2}{*}{\textbf{Deep DA}} 
& Semi-supervised 
& MAML, WheatSeedBelt, SSDA-WheatHead, SSDA-WheatSeg, TDA-YOLO, SCDAL, CDADA \\ \cmidrule(l){2-3}
& Unsupervised 
& MSFF, GAN-DA, DeepDA-Net, U-DA-Net, TriADA, MaxEnt-DA, DAE-DANN, Self-Attn-DA, ADANN, OpenDA, AdaptSegNet, ADVENT, BDL, DFENet, TSAN, DAN, WDGRL, CORAL, Transformer-UDA, MRAN, WPS-DSA, NCADA, TCANet, DDA-Net, TDDA, TST-Net, AMF-FSL, MSUN, MSCN, AMRAN, DJDANs, CLA \\
\bottomrule
\end{tabular}%
}
\end{table*}

\section{Shallow DA Methods}

In agricultural data analysis, manually designed features combined with traditional machine learning algorithms are typically used in shallow domain adaptation (DA) methods. In this section, they are divided into three categories: (1) instance weighting, (2) feature transformation, and (3) classifier adaptation.  

\subsection{Instance Weighting}

In shallow DA, instance weighting is a common approach, especially suitable for agricultural image analysis. Its main idea is to reduce domain differences by adjusting the marginal distributions of source and target samples. Suppose the marginal density functions of the target samples and the source samples are $p_t(x)$ and $p_s(x)$ respectively, then the importance weight $w(x)$ is defined as follows:  

\begin{equation} \label{deqn_ex1a1} w(x) = \frac{p_t(x)}{p_s(x)} \end{equation}

\begin{figure}[ht] \centering \includegraphics[width=1\linewidth]{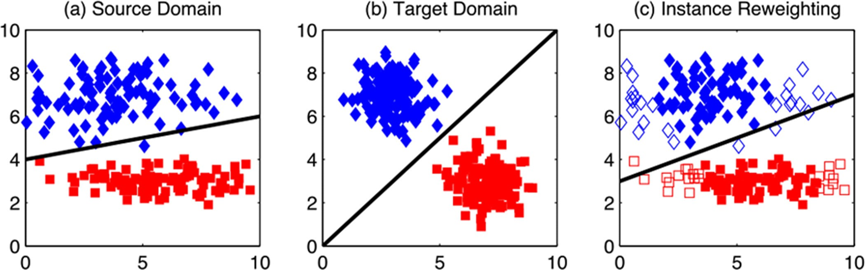} \caption{Domain shift can be alleviated through instance weighting. (a) Source domain following feature matching. (b) Target domain following feature matching. (c) Source domain after applying both joint feature matching and instance weighting, where hollow markers denote samples with low relevance and small assigned weights. Figure adapted from Longet et al.~\cite{5}. } \label{fig3} \end{figure}

In this method, different weights are assigned to source domain samples based on their correlation with the target distribution. For example, crop observation data from geographically close regions may be given higher weights in order to better approximate the target domain. Training on these re-weighted samples can significantly mitigate the distribution gap for classifiers or regressors. As shown in Figure~\ref{fig3}, after introducing instance weighting, the difference between the source domain and the target domain is significantly alleviated.

Molina-Cabanillas et al.~\cite{6} proposed DIAFAN-TL (Domain-Invariant Agricultural Features Adaptation Network for Transfer Learning), which is an instance-weighted learning algorithm for olive tree variety phenology prediction. This method dynamically adjusts the weights of training samples according to the cross-domain correlation, thereby improving the diversity of information and alleviating the problem of sample inconsistency. Yaras et al.~\cite{7} developed Random Histogram Matching (RHM), which models sensor differences and environmental changes as nonlinear pixel transformations, and combines deep neural networks for data augmentation, thereby improving the robustness of satellite images under domain shift.

Cui et al.\cite{8} introduced an iterative weighted active transfer learning (IWATL) approach for hyperspectral image (HSI) classification. By adaptively adjusting the weights of source domain samples through a dual evaluation strategy, their method achieved notable improvements in classification accuracy. Li et al.~\cite{9} introduced a framework known as cost-sensitive self-paced learning (CSSPL), which combines mixed weight regularization to select multi-temporal samples, effectively solving the seasonal and temporal variation problems in agricultural monitoring.

In general, instance-based shallow DA methods usually contain two steps: first, adjust the data distribution by re-weighting or sample selection; then, use the adjusted data to train a stronger classifier. These methods have been shown to be effective in agricultural remote sensing, because agricultural data are often highly heterogeneous. They have shown obvious advantages in transferability, scalability and generalization.

\subsection{Feature Transformation}

Feature-based DA methods map source and target domain data to a shared feature space, with the aim of learning domain-invariant representations to reduce the distribution gap~\cite{10}. Common techniques include subspace alignment, manifold learning, and low-rank representation.

\begin{figure}[ht]
    \centering
    \includegraphics[width=1\linewidth]{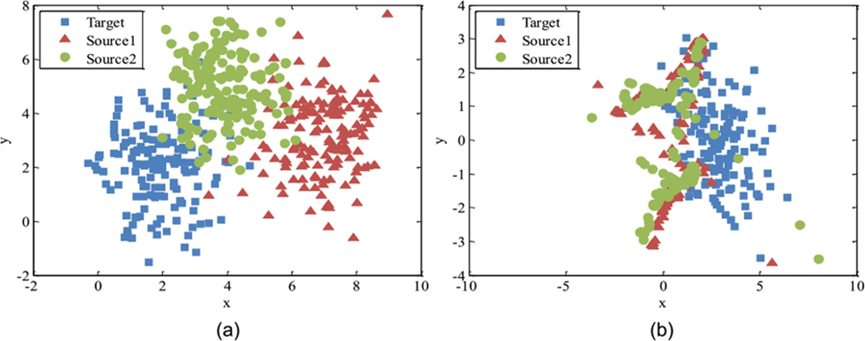}
    \caption{Feature transformation in domain adaptation. Colors denote different domains. (a) The feature distributions of two source domains and one target domain before transformation. (b) The distributions after alignment via feature transformation. Figure adapted from Wang et al.~\cite{11}.}
    \label{fig4}
\end{figure}

As shown in Figure~\ref{fig4}, this type of method first constructs a common space that can retain domain-invariant features, and then trains the classifier on the transformed source domain features and their labels, so that it can make more accurate predictions on the target domain samples.

There are two main types of common implementation strategies: one is \textit{subspace-based projection and reconstruction}, which aligns features thru linear or orthogonal mapping; the other is \textit{re-encoding based on nonlinear transformation}, which uses kernel methods or manifold embedding to model complex relationships.

The core objective is to match the statistical properties of the source domain and target domain features, so as to improve the generalization ability of the model under new conditions. This is especially important in agricultural applications, because seasonal changes, regional differences and sensor differences often cause significant domain shifts.


\subsubsection{Subspace-Based Adaptation}

\begin{figure}
    \centering
    \includegraphics[width=1\linewidth]{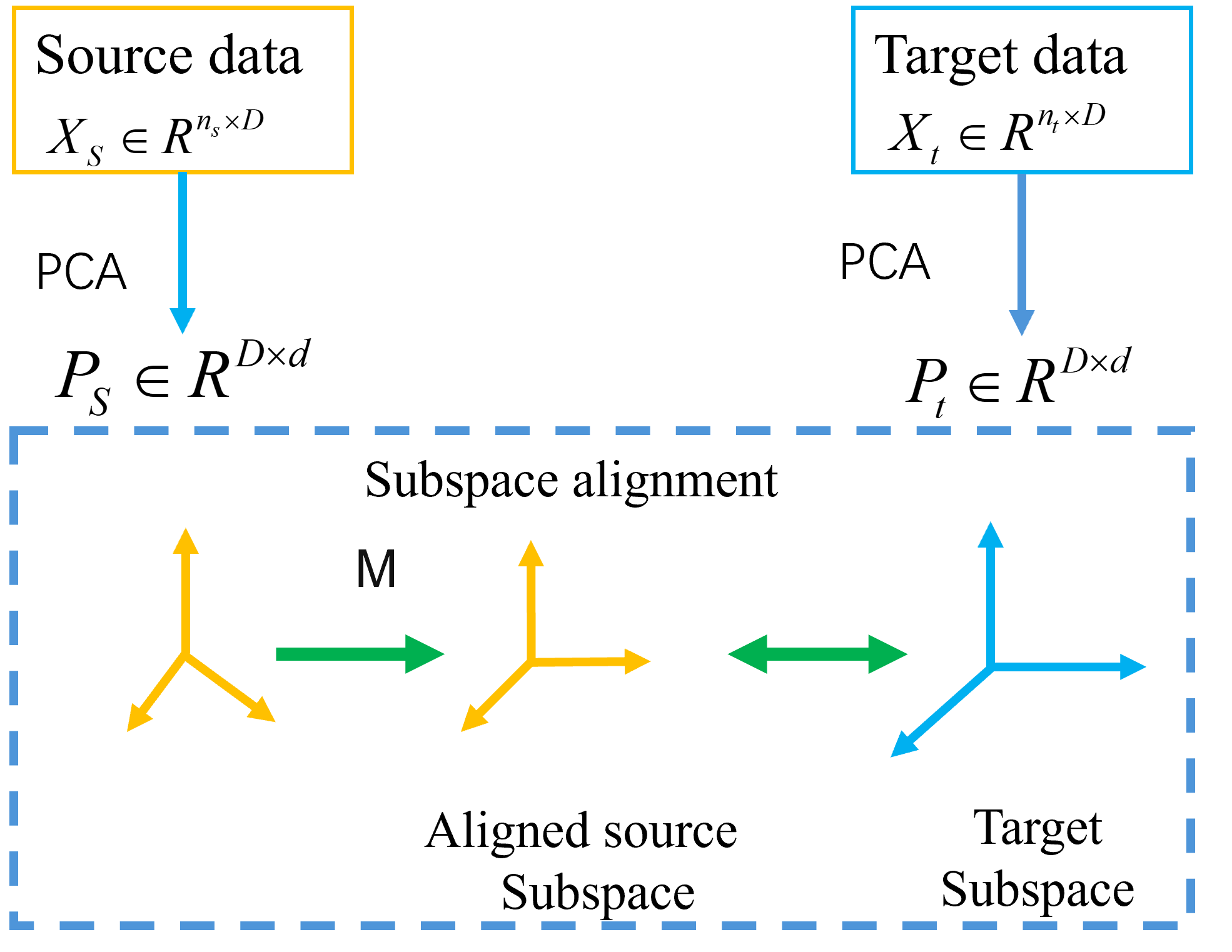}
    \caption{Diagram of Subspace Alignment.}
    \label{fig5}
\end{figure}

In subspace-based domain adaptation (DA), source and target samples are first mapped into their respective subspaces—often obtained through dimensionality reduction or subspace learning—and then aligned to reduce distributional gaps. The Subspace Alignment (SA) method proposed by Fernando et al.~\cite{12} has been extended to multispectral agricultural image classification, where Principal Component Analysis (PCA) is used to align crop spectral feature subspaces across seasons, thereby alleviating domain shifts caused by lighting and phenological variations. The SA process is shown in Figure~\ref{fig5}.  

Let the source domain samples $X_S \in \mathbb{R}^{n_s \times D}$ contain $n_s$ observations with $D$-dimensional features, and the target domain samples $X_T \in \mathbb{R}^{n_t \times D}$ have the same dimensionality.PCA is applied to each domain to obtain principal components, resulting in the projection matrices $P_s$ and $P_t$. A linear transformation matrix $M$ is then learned to perform subspace alignment:  

\begin{equation}
\label{2}
M = \underset{M}{\operatorname*{arg\,min}}\parallel P_s M - P_t \parallel_F^2 = P_s^T P_t
\end{equation}
where $\|\cdot\|_F^2$ refers to the Frobenius norm.  

This method has been used in several agricultural studies. For example, Sun et al.~\cite{13} employed subspace alignment (SA) for cross-view scene classification, where Partial Least Squares (PLS) was utilized to construct a discriminative subspace shared by multiple sensors.
They also proposed Transfer Sparse Subspace Analysis (TSSA)~\cite{14}, which integrates sparse subspace clustering with Maximum Mean Discrepancy (MMD) minimization to preserve self-expressiveness across maize growth stages. Weilandt et al.~\cite{15} proposed the Early Crop Mapping based on Dynamic Clustering Method (ECMDCM), which makes use of time-series NDVI and EVI data for early-stage crop classification.  
 Li et al.~\cite{16} developed CropSTGAN, incorporating a domain mapper to adapt spatial–temporal differences in crop mapping, while Wang et al.~\cite{17} presented MultiCropGAN, which adds an identity loss to preserve key features during label space adaptation, validated on farmland in North America.  

Adversarial learning has also been integrated into subspace alignment. Takahashi et al.~\cite{18} proposed a category-guided feature alignment framework that maps features from both domains into a shared semantic space through adversarial training. Without requiring target labels, this method uses source categories to improve domain-invariant feature learning and enhance generalization across crop types and acquisition conditions.  

A special branch of subspace adaptation can be viewed as feature-invariant methods. The goal of these methods is to find features that remain stable across different domains. Bruzzone et al.~\cite{19} applied multi-objective optimization to choose spatially invariant features for the classification of non-overlapping scenes, with the purpose of balancing discriminability and spatial robustness. Invariant features can also be extracted in Reproducing Kernel Hilbert Spaces (RKHS)~\cite{20}.
Paris et al.~\cite{21} proposed a sensor-driven hierarchical DA framework that makes use of RKHS features, while Yan et al.~\cite{22} introduced a TrAdaBoost variant that incorporates Particle Swarm Optimization (PSO) to achieve optimal subspace selection for cross-domain

To address challenges in hyperspectral image classification—such as label–feature inconsistency and computational inefficiency in similarity learning—Yi~\cite{23} proposed a subspace learning framework based on set similarity metrics, achieving 8\%--15\% higher accuracy and 40\%--60\% shorter training times than traditional methods (e.g., LDA) on datasets such as Indian Pines and Salinas Valley. Banerjee~\cite{24} further developed a hierarchical subspace alignment method that combines a semantic-driven binary tree with Grassmannian manifold alignment to reduce multi-temporal discrepancies caused by atmosphere, sensors, and labeling. Compared with methods such as Global GFK and SA, this approach improves accuracy by 5\%--6\% and significantly reduces misclassification among semantically similar crop categories, providing a scalable solution for dynamic agricultural monitoring and environmental assessment.  


Transformation-based domain adaptation (DA) methods directly change the distribution of data or features to reduce the difference between domains, rather than projecting to the aligned subspace. These methods usually use mapping functions to transform data into a shared representation space, or directly adjust at the data level, so that the source domain and the target domain are more consistent in statistical characteristics.

\begin{figure} \centering \includegraphics[width=1\linewidth]{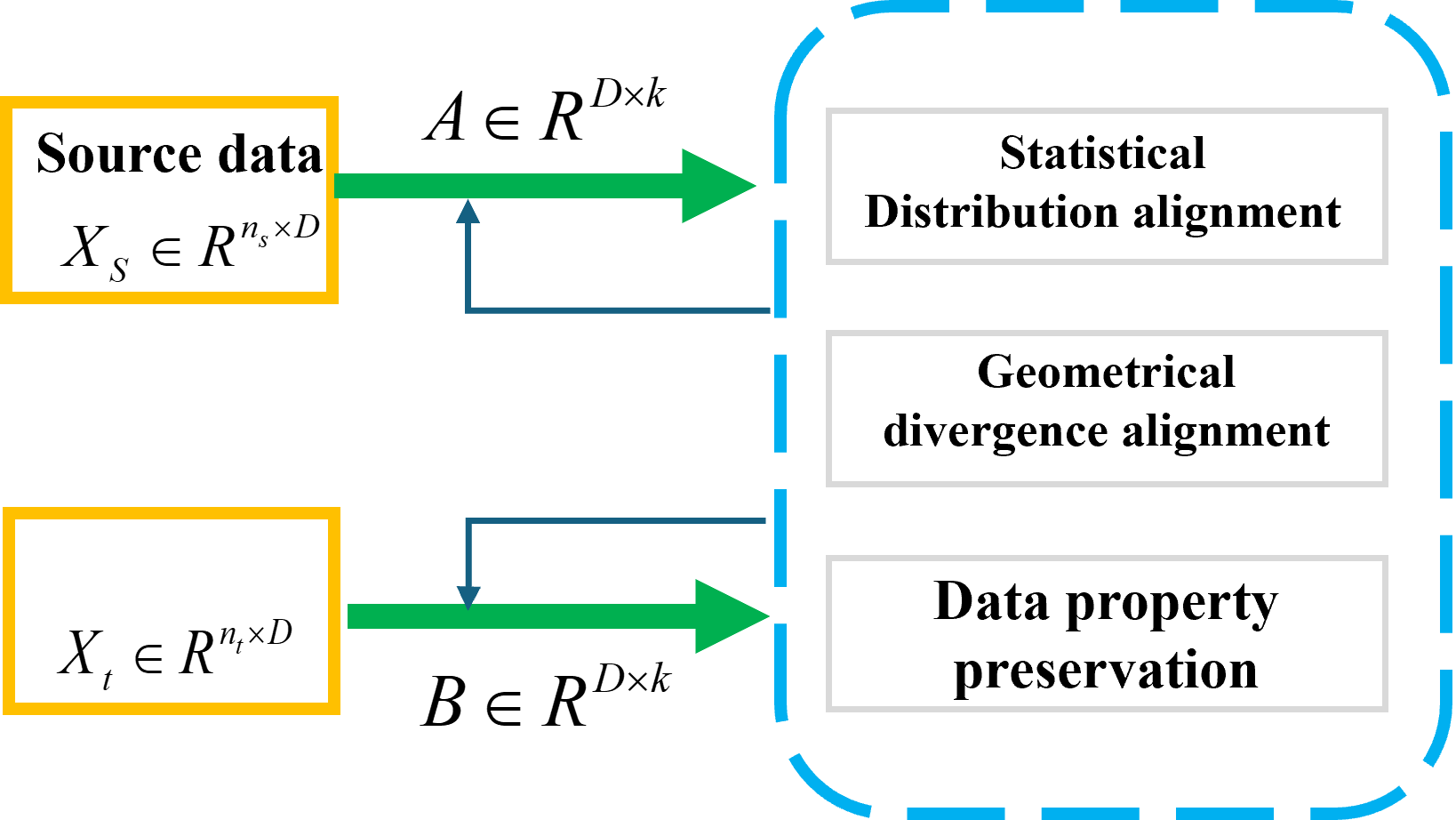} \caption{Schematic diagram of domain adaptation based on transformation.} \label{fig6} \end{figure}

Aptoula et al.~\cite{25} proposed a deep learning-based crop and weed domain adaptive classification framework to solve the problem of distribution inconsistency caused by differences in illumination, soil conditions, and crop phenology. This method maps agricultural remote sensing data to a new feature space, while reducing the statistical and geometric differences across domains and maintaining structural consistency, demonstrating the application value of transformation-based DA. As shown in Figure~\ref{fig6}, this type of strategy has been proven to be effective in agricultural remote sensing and crop monitoring.

These methods often measure the difference between the source domain and the target domain thru divergence measures, including maximum mean discrepancy (MMD), Kullback-Leibler (KL) divergence, and Bregman divergence.Among these measures, MMD is widely applied in agricultural studies because of its robustness in high-dimensional feature spaces. Its definition is as follows:

\begin{equation} \label{3} \text{MMD}(X_s, X_t) = \left\|\frac{1}{n_s} \sum_{\mathbf{x}_i \in X_s} \phi(\mathbf{x}_i) - \frac{1}{n_t} \sum_{\mathbf{x}_j \in X_t} \phi(\mathbf{x}_j) \right\|_F^2 \end{equation}

In this case, $\phi(\cdot)$ is the function that takes data and puts it in the reproducing kernel Hilbert space (RKHS). $X_s$ and $X_t$ represent the data sets of the source domain and the target domain, respectively.

When dealing with nonlinear data such as hyperspectral images, the selection of kernel function is particularly critical: \begin{itemize} \item \textbf{Radial basis function (RBF) kernel}: It is often used for hyperspectral data, which can capture the local nonlinear differences between bands and maintain a balance between accuracy and computational cost.
\item \textbf{Polynomial kernel}: It is suitable for the case where there is correlation between bands, but high-order polynomials are prone to overfitting and increase the amount of calculation.

\item \textbf{Sigmoid kernel}: It is similar to the activation function of neurons in form, but it is rarely used in high-dimensional agricultural data because the training process is unstable and the scalability is limited.
\end{itemize}

In practical applications, the RBF kernel is most frequently adopted for hyperspectral and other complex agricultural datasets because of its robustness in capturing nonlinear structures.
 Pan et al.~\cite{26} proposed transfer component analysis (TCA), which projects the source and target domain data into a shared RKHS and reduces the difference between domains thru MMD. Its objective function is:

\begin{equation} \label{4} \min \operatorname{tr}(W^\top KLK^\top W) + \mu\operatorname{tr}(W^\top W) \end{equation} where $K$ is the kernel matrix, $L$ is the MMD matrix, $W$ is the transformation matrix, and $\mu$ is the regularization parameter. The adaptability of TCA to the changes in the growth stage and the low labeling requirements make it particularly suitable for pest detection, yield estimation and resource allocation.

Canonical correlation analysis (CCA) and its improved methods also provide effective transformation solutions for heterogeneous domain adaptation in agriculture. These methods enhance the robustness and cross-domain generalization ability of the model by modeling domain differences and adjusting the transformation to make it more in line with the characteristics of agricultural data, so as to provide support for scalable and reliable smart agriculture systems.

\subsection{Classifier-Based Adaptation}

Classifier-based DA methods adjust the parameters of the classifier trained on the source domain by using unlabeled or few-labeled target samples. These methods are especially suitable for agricultural scenarios, where environmental conditions change frequently, seasonal variations are significant, and labeled data is often limited.

In agricultural image analysis, domain shift often comes from seasonal differences, crop phenotype differences, and changes in imaging conditions. To address this issue, Bruzzone et al.~\cite{27,28,29,30} proposed a maximum likelihood classifier adaptation scheme that iteratively updates the classifier parameters to match the statistical characteristics of the new season images. On this basis, a Bayesian cascade classifier~\cite{28} was also proposed to improve the stability of crop type identification across regions. Subsequently, Zhong et al. ~\cite{31} combined spectral-spatial guided filtering with classifier adaptive updating to improve the accuracy of crop disease detection under diverse field conditions.

Considering the complexity of cross-domain agricultural data, ensemble learning has gradually become an effective way to enhance robustness. By fusing the predictions of multiple base classifiers, the ensemble method can reduce overfitting and model bias. Wei et al.~\cite{32} proposed a multi-domain adaptive fusion (MDAF) method, which used classifier integration to improve the accuracy of cross-national farmland classification and effectively deal with the problem of spectral differences.

Weighted classifier transfer methods, especially those combined with random forests, have also been widely used in agriculture. Zhang et al.~\cite{33} used the weighted random forest method to integrate agricultural data from different regions, thereby improving the effect of cross-regional crop disease detection. Xu et al.~\cite{34} introduced kernel alignment technology into the extreme learning machine (ELM) framework for crop yield prediction under different climatic conditions, which once again proved the flexibility of classifier-based adaptive methods in agricultural analysis.

To reduce the high cost of agricultural image annotation, semi-supervised learning (SSL) and active learning (AL) methods are increasingly used. Rajan et al.~\cite{35} proposed a binary hierarchical classifier (BHC) to incorporate SSL into cross-year pest monitoring, which realized the continuous improvement of the model. Improvements to the classic domain adaptive support vector machine (DASVM) have also been shown to improve the performance of cross-sensor crop classification when there are very few labeled samples in the target domain~\cite{36}.

\begin{figure} \centering \includegraphics[width=1\linewidth]{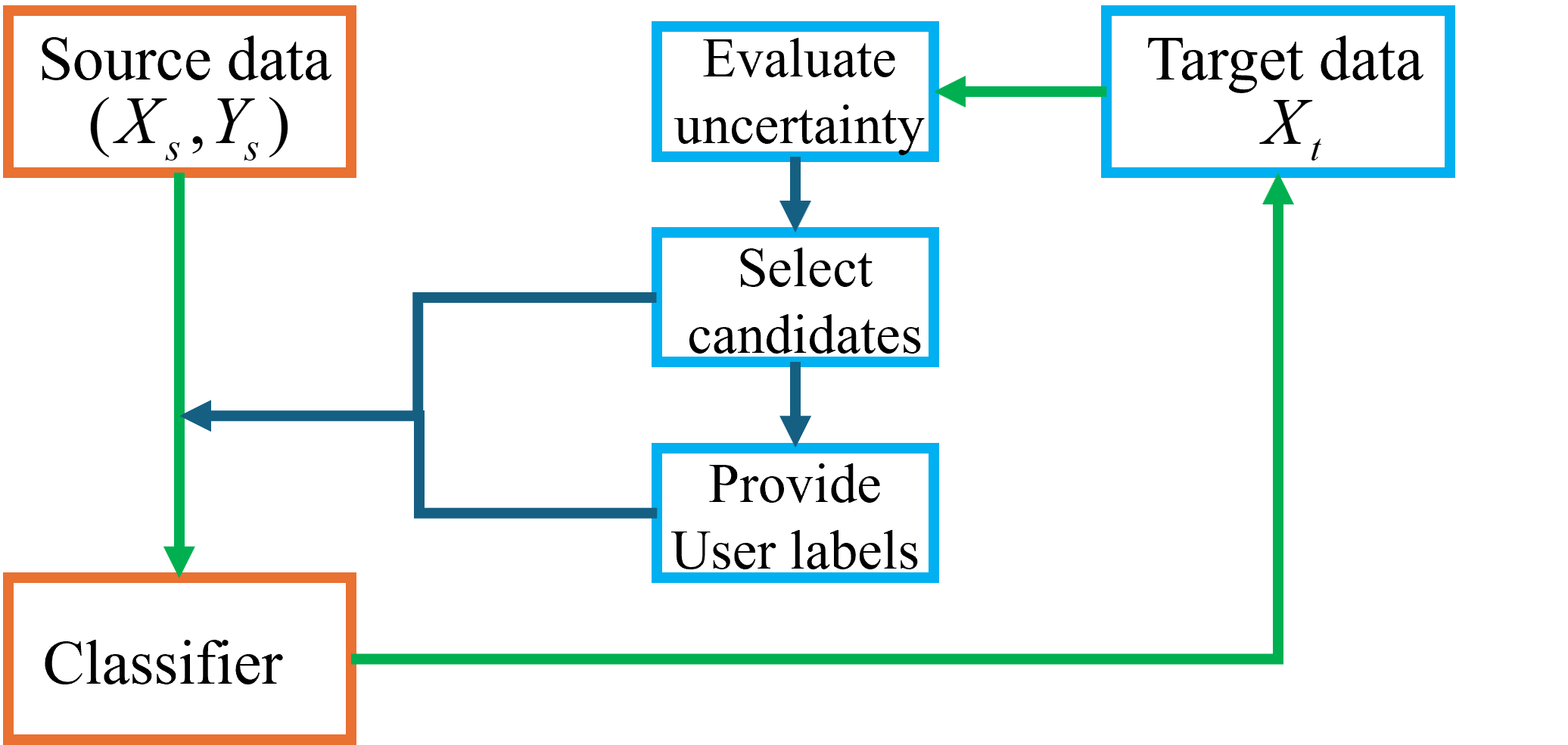} \caption{Flowchart of domain adaptation based on active learning (AL).} \label{fig7} \end{figure}

As shown in Figure~\ref{fig7}, a typical AL-based DA process includes: first, training a classifier with source domain data and using it to automatically label target domain samples; then, manually labeling the samples with the highest prediction uncertainty and adding them to the training set, so as to iteratively improve the classifier performance. This strategy is particularly effective in precision agriculture because it minimizes the need for labeling. Deng et al.~\cite{37} proposed the active multi-kernel domain adaptation (AMKDA) method, which combines active learning (AL) with multi-kernel learning, greatly reducing the labeling requirements of hyperspectral data. An active learning (AL) approach that employs standard deviation (SD) to identify informative samples was presented by Kalita et al.\cite{38}. In contrast, Saboori et al.\cite{39} introduced active multi-kernel Fredholm learning (AMKFL), which incorporates Fredholm regularization to guide labeling and enhance classification accuracy.

New classifier designs have also been proposed to address the shape and scale variation problems in agricultural images. Izquierdo-Verdiguier et al.~\cite{40} developed a virtual support vector (VSV) classifier, which introduced rotation and scale invariance for UAV farmland classification, and improved the classification accuracy by up to 12\%. Wang et al.~\cite{41} proposed EasyTL, a non-parametric transfer method for edge devices, which can realize real-time crop state recognition and maintain good adaptability under limited resources.

\subsection{Summary of Shallow DA Methods}

Shallow DA methods generally rely on traditional machine learning algorithms and handcrafted features to reduce the gap between the source domain and the target domain.
 They can be generally divided into three categories: \textit{instance weighting}, \textit{feature transformation} and \textit{classifier-based adaptation}. Instance weighting reassigns the sample weights to match the target distribution; feature transformation projects the source and target domains into a shared space to achieve feature alignment; classifier adaptation improves the target domain performance by updating or regularizing the classifier parameters.

Shallow DA methods have the advantages of simple structure, low cost, and low computational requirements, and do not rely on large-scale computing resources. This makes it applicable in agricultural scenarios, especially suitable for deployment in edge devices or areas with limited infrastructure.

However, these methods have obvious limitations when facing high-dimensional, complex or large-scale agricultural datasets. Their limited feature extraction capabilities make it difficult to characterize nonlinear relationships and contextual patterns. When the domain differences are significant (such as cross-regional crop mapping or multi-season disease detection), shallow methods often fail to capture deeper semantic associations, resulting in a decline in generalization ability.

These limitations have driven the development of DA methods based on deep learning. Deep models are able to learn rich multi-level features and characterize complex domain patterns, providing more robust and scalable cross-domain agricultural image analysis solutions. At the same time, they also show better adaptability in dealing with diversified and dynamically changing data conditions.

\section{Deep Domain Adaptation (DA) Methods}

With the rapid development of deep learning, domain adaptation (DA) methods based on deep neural networks have shown significant advantages in agricultural image analysis. These methods can automatically learn domain-invariant representations thru end-to-end feature extraction and distribution alignment, thereby alleviating the domain shift caused by sensor differences, environmental conditions, and changes in crop growth stages. According to the amount of labeled data in the target domain, deep DA methods can be generally divided into three categories: supervised, semi-supervised, and unsupervised. This section will introduce the basic principles and the latest application progress in agriculture.

\subsection{Supervised Deep DA}

In supervised deep domain adaptation (DA), it is assumed that only a small set of labeled samples is available in the target domain. These methods optimize the feature extractor and classifier simultaneously by minimizing the weighted combination of source domain classification loss and cross-domain alignment loss. The core idea is to introduce the supervision of target domain labels in the domain alignment process, so as to improve the cross-domain generalization ability. A typical supervised deep DA framework consists of three parts: a feature extractor, a classifier, and a domain difference measurement module. Its joint loss function is defined as follows:

\begin{equation} \label{5} L = L_{cls}(X_s, Y_s) + \lambda \cdot L_{align}(X_s, X_t) \end{equation}

Here, $L_{cls}(X_s, Y_s)$ is defined as the classification loss on the source domain,$L_{align}(X_s, X_t)$ measures the domain difference (such as MMD, adversarial loss or covariance alignment), and $\lambda$ is the weighting parameter.

In agricultural applications, supervised deep DA has been extensively applied to cross-sensor data fusion and multi-temporal crop monitoring. For example, Takahashi et al.~\cite{18} proposed a category-aware adversarial learning framework, which uses the source domain category label to guide the feature space alignment, thereby improving the generalization ability on the agricultural data set collected on the heterogeneous imaging platform.

\begin{figure} \centering \includegraphics[width=1\linewidth]{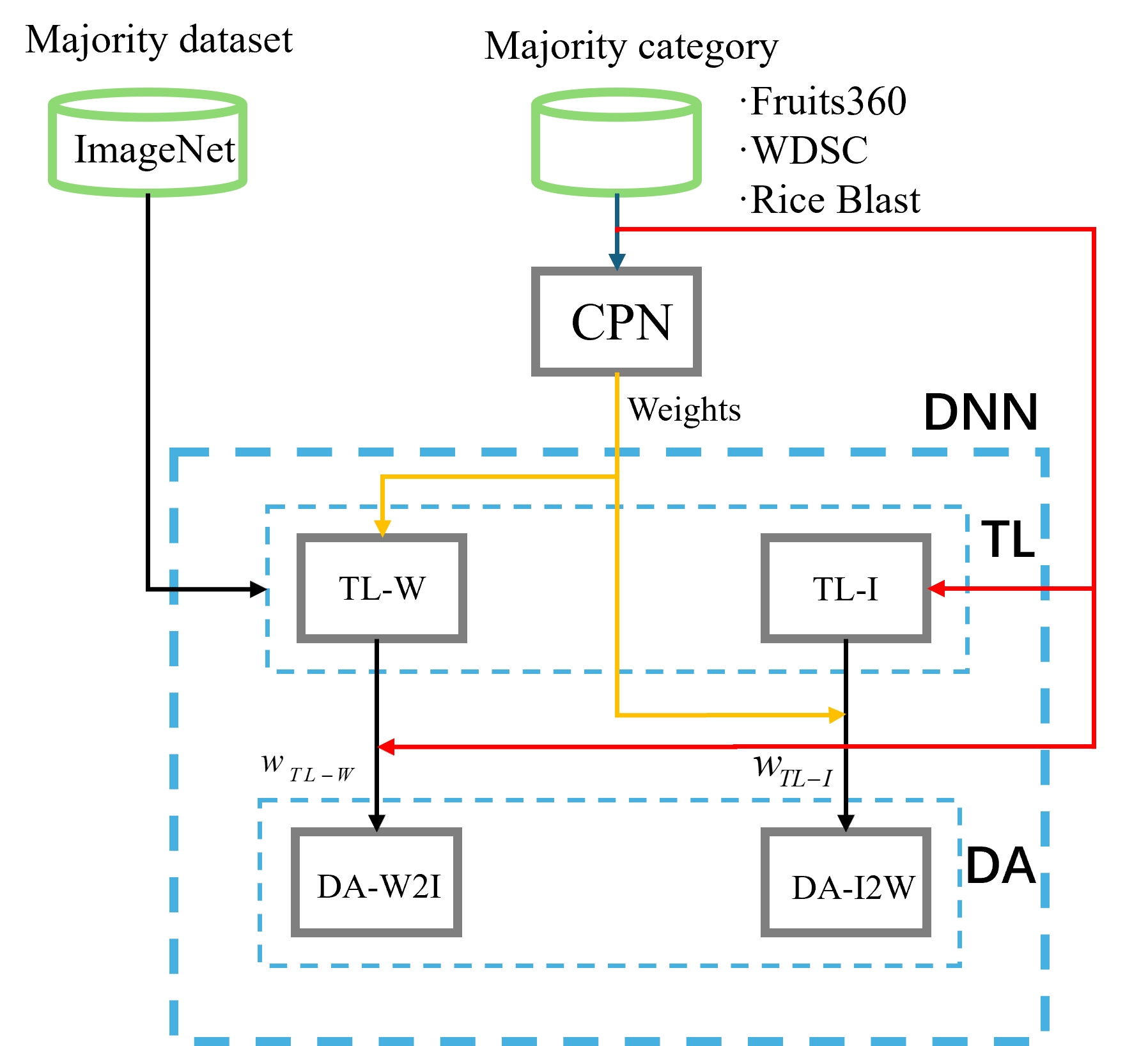} \caption{The structure diagram of Takahashi et al.'s method~\cite{18}.} \label{fig8} \end{figure}

As illustrated in Figure~\ref{fig8}, the proposed approach is structured in three stages: image generation, transfer learning, and domain adaptation with a deep neural network (DNN). First, the category propagation network (CPN) generates category maps from the target images to simulate the real distribution of agricultural data. Then, transfer learning is performed on the real images ($I$) and the synthetic images ($W$) to obtain two models, TL-I and TL-W. Finally, the bidirectional DA modules (DA-W2I and DA-I2W) are used to realize the image-level and feature-level adaptation from synthetic to real and from real to synthetic, respectively. The system consists of five parts, namely CPN, TL-I, TL-W, DA-W2I and DA-I2W, which work together to enhance the generalization ability of agricultural tasks in the target domain under limited annotation conditions~\cite{45}. Similarly, Espejo-García et al.~\cite{43} combined agricultural transfer learning with generative adversarial networks (GANs) to synthesize labeled images to alleviate the problem of label scarcity.

Supervised convolutional neural networks (CNNs) have also been successfully applied to segmentation tasks in agricultural remote sensing. Lu et al.~\cite{46} introduced the attention mechanism to highlight the key spatial features and combined the feature fusion module to improve the segmentation accuracy of farmland boundary extraction. However, low-resolution feature maps tend to cause boundary blurring during upsampling. To solve this problem, Li et al.~\cite{47} proposed a pixel-by-pixel context modeling method and combined it with post-processing to improve the clarity of the boundary.

To address the scale variation problem, Shang et al.~\cite{48} developed a multi-scale target extraction framework that combines the attention mechanism, which can realize adaptive segmentation of agricultural objects of different sizes. Zhang et al.~\cite{49} reformulated farmland extraction as an edge detection task and designed a high-resolution boundary optimization network (HBRNet), which significantly improved the boundary accuracy.

Despite the progress made, supervised deep DA is still limited by the high cost of obtaining labeled data in the target domain in practical agricultural applications. For example, in the Cityscapes dataset~\cite{50}, it takes about 1.5 hours of manual work to label an image. This challenge highlights the importance of semi-supervised and unsupervised DA methods, which can make full use of a large amount of unlabeled data, reduce the cost of labeling, and still maintain the cross-domain adaptability in real agricultural scenarios.


\subsection{Semi-Supervised Deep Domain Adaptation (DA)}

Semi-supervised deep DA methods combine fully labeled source domain data with a small number of labeled or pseudo-labeled target domain samples to achieve cross-domain model training and optimization.
Common strategies include self-training, co-training, and mixed supervision. Its main advantage is that it can achieve a balance between labeling cost and model performance, which is especially suitable for agricultural scenarios where target domain labeling data is scarce.

Self-training with pseudo-label generation forms the basis of many semi-supervised DA. In recent years, diffusion models have been introduced into this paradigm and have shown good results in agricultural applications. Ghanbari et al.~\cite{51} proposed a framework combining diffusion model and meta-learning for wheat ear segmentation under small sample conditions. The method only needs three manually labeled wheat images and a set of unlabeled video frames to generate a large-scale pseudo-labeled data set thru the probabilistic diffusion process.

\begin{figure} \centering \includegraphics[width=1\linewidth]{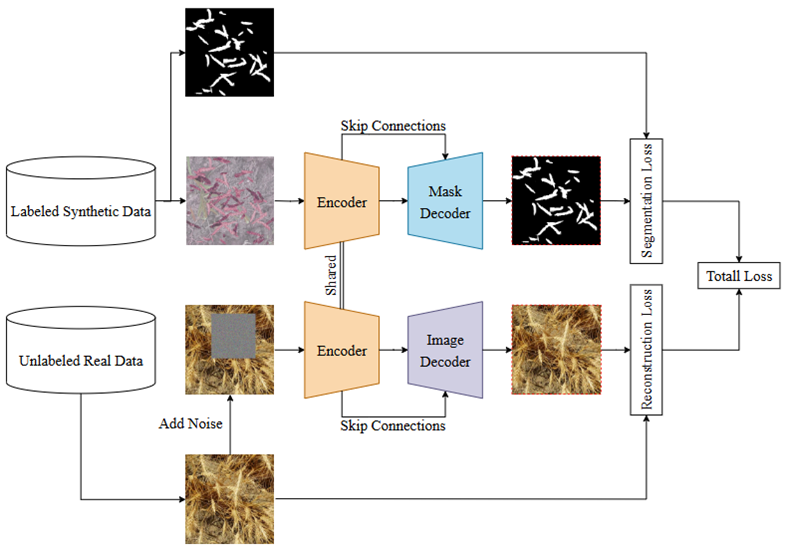} \caption{Schematic diagram of the model structure of Ghanbari method~\cite{51}.} \label{fig9} \end{figure}

The framework adopts a dual-branch encoder-decoder design to achieve feature alignment and maintain cross-domain semantic consistency (see Figure~\ref{fig9}). The Dice score on the external farmland dataset from multiple countries reached 63.5

In this architecture, the encoder learns shared features from real and synthetic images; the mask decoder outputs segmentation results; and the image decoder reconstructs real images, thereby driving the encoder to better adapt to real-domain features. Experiments show that the Dice score on the internal dataset reaches 81\%, and the Dice score on the external dataset from 18 plots in five countries reaches 64.8\%. With the help of diffusion models, the framework can utilize large-scale unlabeled data, showing significant value in image processing~\cite{52,53} and precision agriculture.

In terms of wheat disease detection, Najafian et al.~\cite{54} constructed the WheatSeedBelt dataset and proposed a pipeline based on pre-trained models and semi-supervised fine-tuning, which achieved high-precision classification of Fusarium head blight kernels (FDK). Subsequently, they proposed a semi-self-supervised method for wheat ear detection~\cite{55}, which used a small number of labeled images and videos to generate pseudo-labeled datasets, and further improved thru domain adaptation, with mAP reaching 0.827. At the same time, they designed a hybrid method that combines synthetic data with multi-domain adaptation to achieve wheat ear segmentation~\cite{56}. The method only needs two labeled samples, and the Dice score on the internal data is 0.89. After fine-tuning, it is increased to 0.91 on diverse external data, showing a strong generalization ability in the low-labeled scene.

In vegetable detection, Yang et al.~\cite{44} proposed the TDA-YOLO model, which is based on YOLOv5 and optimized for dense tomato detection, and can adapt to different lighting conditions. It uses neural style transfer to generate pseudo-datasets to reduce the domain gap, and combines semi-supervised learning and knowledge distillation to improve the adaptability of the target domain. At the same time, it introduces the lightweight FasterNet backbone network to speed up the reasoning speed. Zhu et al.~\cite{57} proposed a semi-supervised center discriminative adversarial learning (SCDAL) method, which enhances the cross-domain discrimination ability thru adversarial learning guided by center loss, and is used for aviation scene classification. Teng et al.~\cite{78} proposed a classifier-constrained deep adversarial domain adaptation (CDADA) framework, which combines the maximum classifier discrepancy (MCD) with deep convolutional neural networks (DCNNs) to achieve semi-supervised remote sensing cross-domain classification.

In summary, semi-supervised domain adaptation uses a small amount of labeled data to guide feature alignment, which reduces the labeling requirements and improves the robustness in heterogeneous agricultural scenarios. This type of method also supports multi-modal data fusion and stage-aware crop monitoring, which is of great significance for building a scalable precision agriculture system.

\subsection{Unsupervised Deep Domain Adaptation (DA)}

Unsupervised Deep DA makes full use of the labeled source domain data and unlabeled target domain samples to learn domain-invariant representations thru feature space transformation or generative models~\cite{124,125}. Common methods can be divided into three categories: adversarial-based, discrepancy-based, and self-supervised learning. These methods have shown significant value in application scenarios with scarce labels, such as agricultural robots and remote sensing mapping~\cite{139}.

\subsubsection{Adversarial Training}

In unsupervised domain adaptation (UDA), adversarial training, especially \textbf{Generative Adversarial Networks (GANs)}, is widely used to align the distributions of source and target domains. The most representative methods are \textbf{Domain-Adversarial Neural Network (DANN)} and \textbf{Gradient Reversal Layer (GRL)}, whose concepts are derived from the GAN framework.

GANs are a type of deep generative model~\cite{58} and are often used for image synthesis tasks. Its training depends on the representative data sets of the target domain, such as CelebA (face image)~\cite{59}, MNIST (handwritten digits)~\cite{60}, LSUN (indoor scene)~\cite{61}, or CIFAR-10~\cite{62} and ImageNet~\cite{63} and other general image data sets. After training, the generator can generate images that are almost indistinguishable from real samples.

\begin{figure}[ht] \centering \includegraphics[width=1\linewidth]{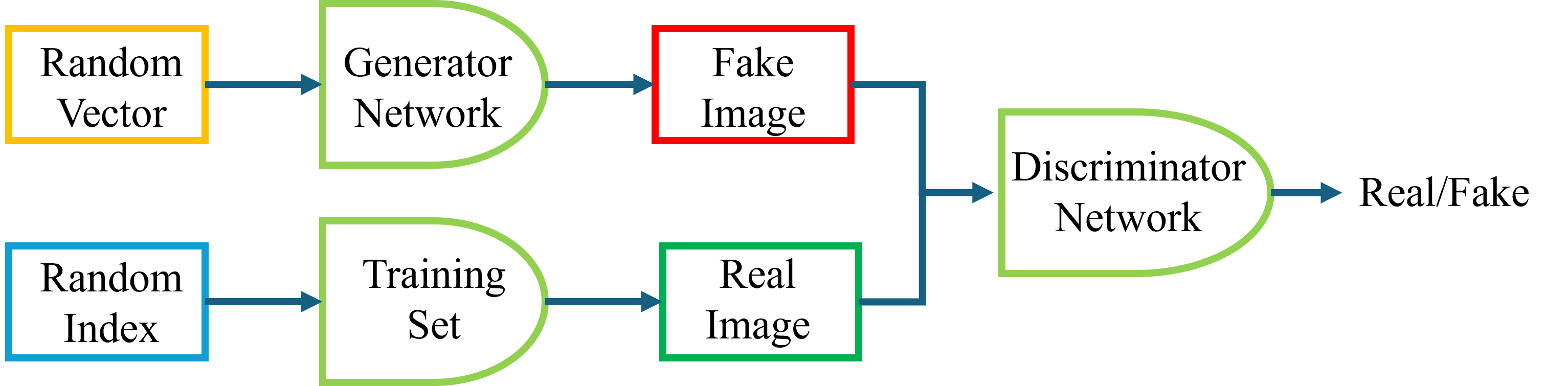} \caption{Schematic diagram of GAN adversarial training principle. The generator continually tries to produce realistic samples, while the discriminator learns to tell real from fake data. } \label{fig10} \end{figure}

As shown in Figure~\ref{fig10}, a typical GAN consists of two adversarial components~\cite{80}: \begin{itemize} \item \textbf{Generator}: Its input is a noise vector (usually sampled from a uniform distribution or a Gaussian distribution), and its goal is to generate samples that are as close as possible to the real data; \item \textbf{Discriminator}: Its input may be a real sample or a pseudo sample synthesized by the generator, and its role is to judge the authenticity of the input data.
\end{itemize}

During the training process, the generator and discriminator compete with each other and jointly complete a \textit{minimax optimization process}. The goal of the generator is to fool the discriminator, while the discriminator is constantly learning how to avoid being fooled. Goodfellow et al.~\cite{58} formalized this process as:

\begin{align} \label{6} \min_G \max_D V(D, G) &= \mathbb{E}_{x \sim P_{\mathrm{data}}}[\log D(x)] \notag\\ &\quad + \mathbb{E}_{z \sim P_z}[\log(1 - D(G(z)))] \end{align}

where \( x \sim P_{\text{data}}(x) \) represents the sample from the real data distribution, \( z \sim P_z(z) \) is the noise sampled from the prior distribution, $D(x; \theta_d)$ is defined as the probability given by the discriminator that the sample $x$ originates from the real dataset, while $G(z; \theta_g)$ denotes the sample produced by the generator from the noise input $z$.

The goal of the discriminator is to \textit{maximize} the accuracy of distinguishing between true and false samples, while the generator \textit{minimizes} the probability of being identified by the discriminator. When $D(G(z)) = 0$, the discriminator completely identifies the fake sample; when $D(G(z)) = 1$, the generator successfully deceives the discriminator.

This adversarial mechanism continuously improves the performance of the generator during the training process, so that it can eventually synthesize highly realistic samples.
 GANs have shown great adaptability, especially when there are obvious visual differences between the source domain and the target domain (such as cross-sensor style migration in agricultural remote sensing).

In adversarial training, \textbf{GAN-like methods} (such as CycleGAN) and \textbf{discriminative adversarial networks} (such as DANN) have different objectives and different applicable scenarios. The former focuses more on synthesizing data samples and is suitable for style transfer and image synthesis; the latter focuses on feature distribution alignment and is often used for feature representation learning and domain adaptation.

\begin{figure}[ht] \centering \includegraphics[width=1\linewidth]{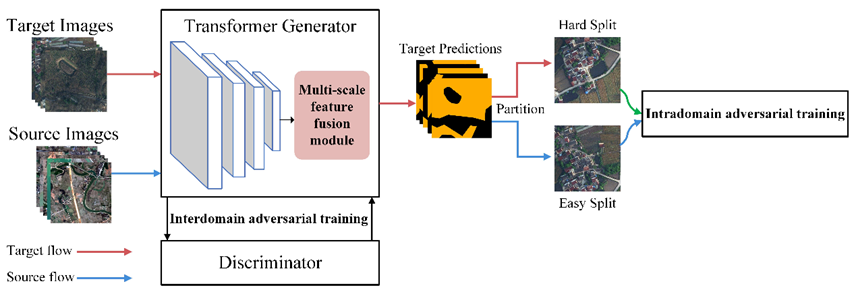} \caption{The multi-scale cross-domain adversarial training and feature fusion framework proposed by Zhang~\cite{65}.} \label{fig11} \end{figure}

In agricultural applications, GANs have been widely used for cross-sensor image style transfer. For example, CycleGAN is used to convert the style of remote sensing images in the source domain to the style of the target sensor, thereby alleviating the domain shift~\cite{64}. Zhang~\cite{65} proposed an unsupervised adversarial domain adaptation-based farmland extraction method (see Figure~\ref{fig11}), and introduced a multi-scale feature fusion (MSFF) module to adapt to data with different spatial resolutions, which improved the learning effect of domain-invariant features.

An end-to-end GAN-based domain adaptation (GAN-DA) framework for land-cover classification with multi-source images was introduced by Ji et al.~\cite{66}. Tasar et al.~\cite{67} designed a GAN-DA pipeline suitable for high-resolution satellite images and extended it to multi-source, multi-objective, and lifelong learning scenarios. Wittich et al.~\cite{68} proposed Color Map GAN for appearance consistency adaptation in aerial image classification. Liu et al.~\cite{69} designed a feature extractor based on GAN to improve the cross-domain accuracy of agricultural scene classification by aligning the distribution of the source domain and the target domain.

\begin{figure}[ht] \centering \includegraphics[width=1\linewidth]{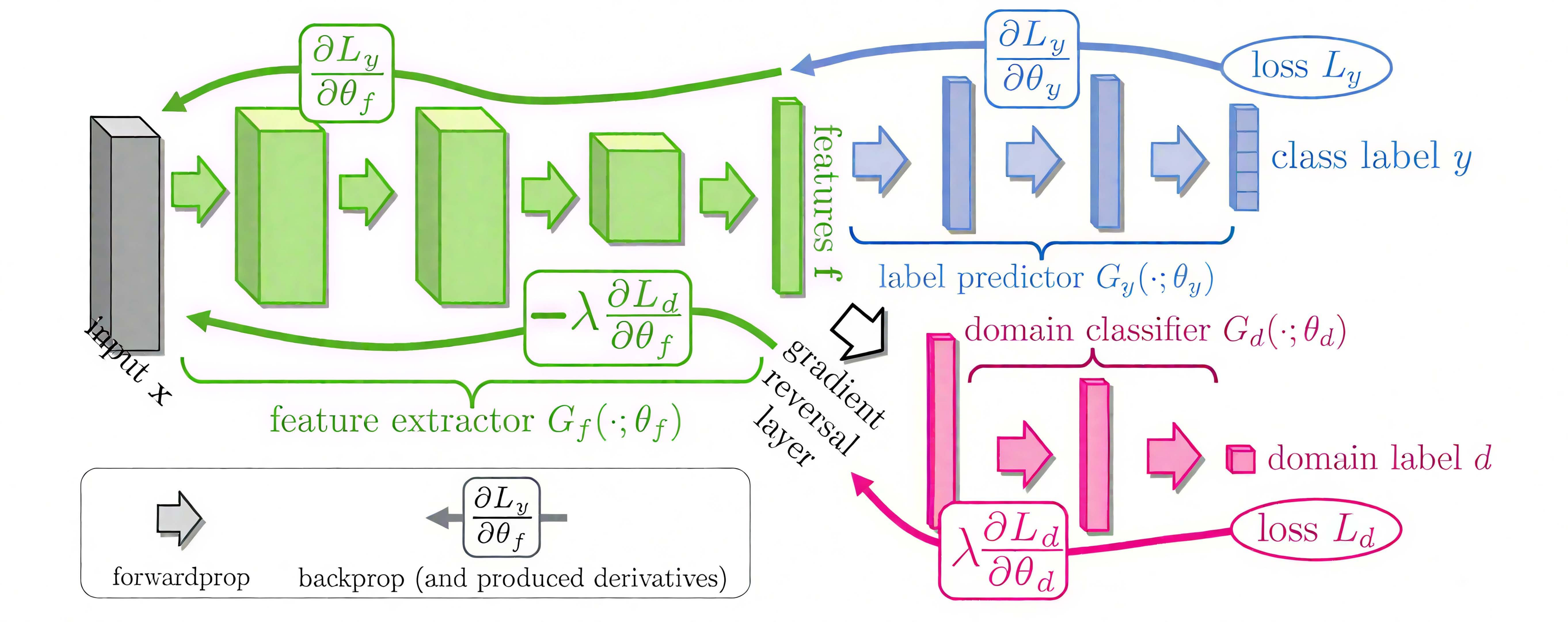} \caption{The basic structure of DANN's cross-domain feature alignment.} \label{fig12} \end{figure}

In contrast, \textbf{discriminative adversarial networks} (such as DANN, see Figure~\ref{fig12}) usually introduce a domain classifier to distinguish the features of the source domain and the target domain. During training, the gradient reversal layer is used to achieve back propagation, so as to implicitly complete the feature alignment. These methods are widely used in agriculture for domain-invariant feature extraction and alignment.

For example, a triple adversarial domain adaptation (TriADA) framework for pixel-level classification of high-resolution remote sensing images was proposed by Yan et al.~\cite{70}. Wang et al.~\cite{71} proposed an adversarial domain adaptation (ADDA) method for cross-platform calibration between hyperspectral imaging devices, which achieved a 21.5\% reduction in the RMSE of maize relative water content (RWC) prediction.

The use of DANN in agricultural image analysis has also been further investigated in other studies. For example, Bejiga et al.~\cite{72} applied DANN to large-scale land cover classification; A multi-source DA approach based on min-max entropy optimization was proposed by Rahhal et al.\cite{73}, whereas Elshamli et al.\cite{74} employed denoising autoencoders (DAEs) in combination with DANN for adapting multi-temporal and multi-spatial data. Mauro et al.~\cite{75} introduced a self-attention module to mitigate domain differences in multi-temporal land cover mapping; Mateo et al.~\cite{76} proposed a cross-sensor adversarial DA method for cloud detection; Ma et al.~\cite{77} proposed an adaptive DANN (ADANN) to align the meteorological characteristics and vegetation index of the agricultural area in the United States, achieving a 34.1\% reduction in the RMSE of yield prediction.

In summary, adversarial-based DA strategies, including hybrid GAN-DANN frameworks, have been extensively applied to remote sensing tasks such as scene classification~\cite{64}, crop classification~\cite{79}, road extraction~\cite{80}, and cloud detection~\cite{81}. Nevertheless, fixed distance metrics (e.g., MMD) may be insufficient for diverse agricultural image datasets. Recent studies have further optimized adversarial DA: Hoffman et al.~\cite{82} first applied adversarial training to pixel-level domain adaptation; Tsai et al.~\cite{83} proposed AdaptSegNet to reduce the domain shift of the segmentation output layer; Vu et al.~\cite{84} proposed ADVENT, which introduced entropy-based adversarial loss into segmentation output; Li et al.~\cite{85} proposed bidirectional learning (BDL) to improve bidirectional feature alignment; Zhang et al.~\cite{86} designed DFENet, which introduced channel and spatial dependency modeling; Zheng et al.~\cite{87} proposed TSAN, a two-stage adaptive network designed for multi-objective remote sensing scene classification; Rahhal et al.\cite{73} further explored multi-source adversarial adaptation through entropy-based alignment, while Makkar et al.~\cite{88} applied adversarial learning to derive geospatial features that are discriminative for the target domain; Saito et al.~\cite{89} proposed an open-set adversarial DA framework, which explicitly distinguishes between known and unknown classes in training, effectively supporting open-set classification.


\subsubsection{Difference-based Alignment Method}

\begin{figure}[ht] \centering \includegraphics[width=1\linewidth]{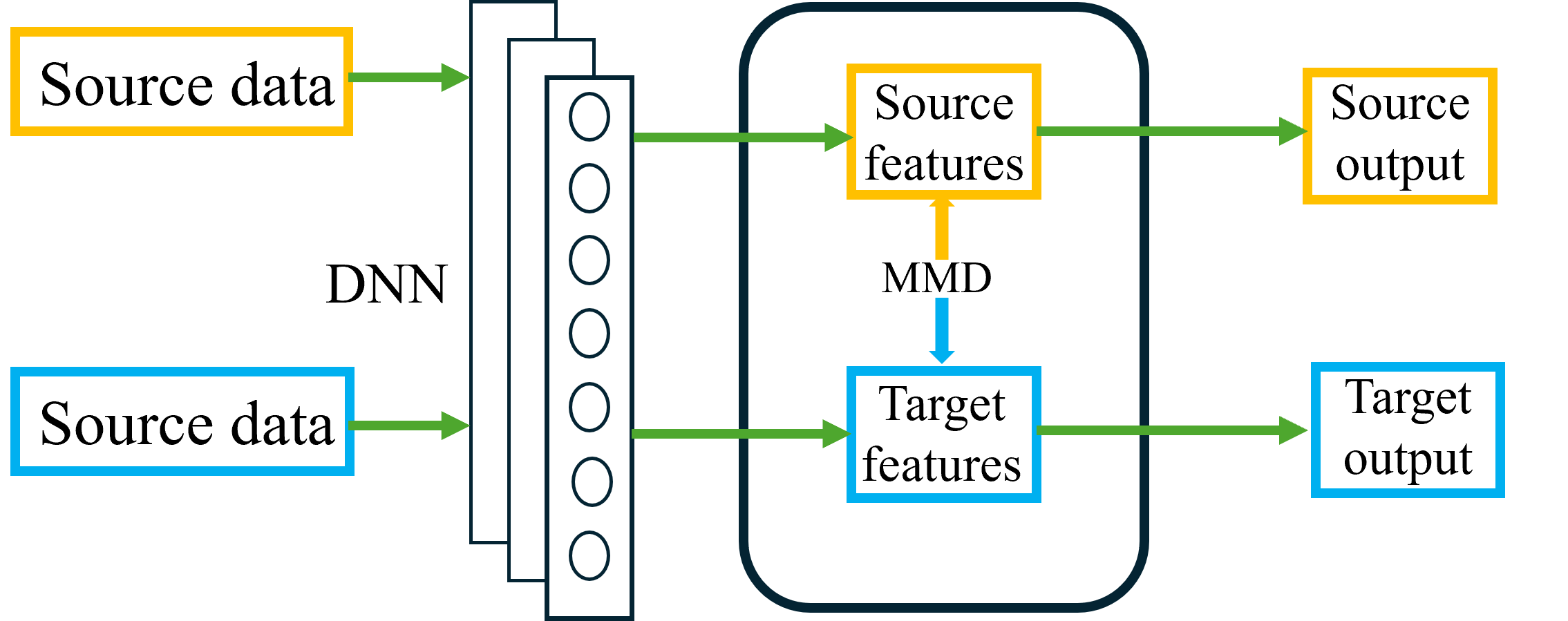} \caption{Flow chart of the difference-based deep DA method.} \label{fig13} \end{figure}

The deep domain adaptation (DA) methods based on discrepancy aim to reduce the marginal or conditional distribution discrepancy between the source and target domains by introducing distribution alignment metrics (such as maximum mean discrepancy MMD) into the deep neural network structure~\cite{90}. These methods usually add additional alignment layers to the network to learn representations that are both task-related and domain-invariant, as shown in Figure~\ref{fig13}.

Long et al.~\cite{91} first proposed the deep adaptation network (DAN), which extends the domain adaptation capability on the basis of the convolutional neural network and is realized by aligning the distribution of cross-domain learning features. Baktashmotlagh et al.~\cite{92} proposed a domain-invariant projection method using MMD, while Shen et al.~\cite{93} introduced a representation learning framework based on Wasserstein distance (WDGRL). Sun et al.~\cite{94} extended the CORAL framework to Deep CORAL, which aligns the activation correlations in the deep layer thru nonlinear transformation.

In the agricultural field, Ferreira~\cite{42} proposed an unsupervised domain adaptation method (UDA-SegFormer) based on visual Transformer, which can improve the detection effect of crop rows and gaps between different farms. By aligning domain-specific features and optimizing pseudo-labels, the method shows good generalization performance in complex environments such as shading and curved rows, and exceeds CNN models (such as PSPNet and DeepLabV3+). This is the first time that Transformer has been introduced into agriculture for UDA, providing an efficient and low-label demand solution for crop monitoring.

Guo et al.~\cite{144} proposed a scheme based on mask image consistency and multi-granularity alignment to improve the domain adaptive ability in plant disease detection. This method effectively reduces the difference between the source domain and the target domain, and achieves the best performance in cross-domain robustness. Gao et al.~\cite{145} constructed a transfer learning framework combining ground images and UAV images for weed segmentation tasks. This method significantly reduces the difference between the aerial and ground perspectives, thus achieving higher precision weed mapping in precision agriculture. Fu et al.~\cite{146} proposed a multi-domain data augmentation method and applied it to crop disease detection in combination with a test-time adaptation strategy. Even in the absence of source domain data, the method can still maintain good generalization performance in the new environment.

Takahashi et al.~\cite{18} improved the performance of agricultural image recognition by aligning the feature space at the category level. Based on this, Long et al.~\cite{95} proposed the multi-representation adaptation network (MRAN)~\cite{96} and the deep subdomain adaptation network (DSAN)~\cite{97}, which further alleviated the cross-domain difference problem thru subdomain division and multi-representation learning.

In high-resolution satellite (HRS) images, there are often differences in scale, resolution, and spectral characteristics between different regions~\cite{98}. To solve this problem, Zhu et al.~\cite{99} proposed a weakly pseudo-supervised decorrelated subspace adaptation (WPS-DSA) network for land use classification, which effectively dealt with the feature distribution shift in space and time.

\begin{figure}[ht] \centering \includegraphics[width=1\linewidth]{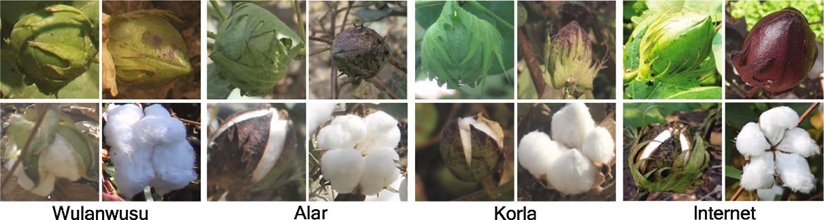} \caption{Example images of the in-field cotton boll (IFCB) dataset.} \label{fig14} \end{figure}

In the task of cotton boll recognition, Li~\cite{100} proposed an unsupervised DA method (NCADA) based on neighborhood component analysis, and constructed the first field cotton boll image benchmark dataset, as shown in Figure~\ref{fig14}. The results show that NCADA surpasses the traditional classification method and maintains a high accuracy under the real distribution change, which provides a new direction for automatic phenotypic analysis.

In hyperspectral image (HSI) domain adaptation, Garea et al.~\cite{101} proposed TCANet. It embeds transfer component analysis (TCA) into deep networks to achieve cross-domain feature alignment. Ian et al.~\cite{58} developed DDA-Net for unsupervised knowledge transfer. A two-stage deep domain adaptation (TDDA) framework was proposed by Li et al.~\cite{103}, where the first stage reduces domain discrepancy via MMD-based deep embedding, and the second stage applies a spatial–spectral dual network to learn discriminative features. Zhang et al.~\cite{104} proposed TST-Net, which combines graph convolutional network (GCN) and convolutional layer to simultaneously encode topological and semantic information. Wang et al.~\cite{105} designed a GNN-based DA model, which incorporates CORAL to achieve joint alignment of domain and category. Liang et al.~\cite{106} proposed AMF-FSL, an attention-driven few-shot learning framework that combines target category alignment, domain attention, and multi-source fusion.

In plant pathology, Yan et al.~\cite{107} proposed a hybrid subdomain alignment strategy for cross-species disease severity identification. Fuentes et al.~\cite{108} built a domain adaptive classification model for tomato diseases and considered environmental changes. However, these methods are still not sufficient to fully cope with the complexity of cross-species diagnosis. To this end, Wu~\cite{109} proposed the MSUN network, a multi-representation subdomain alignment method with uncertainty regularization. This method effectively solves the key problems of intra-domain ambiguity, inter-class overlap and cross-domain heterogeneity in plant disease identification.

Othman et al.~\cite{110} introduced a domain adaptation (DA) network designed for cross-scene classification, which adopted a two-stage training strategy combining pre-training and fine-tuning to achieve cross-domain feature alignment while ensuring the classification accuracy of source domain samples, and maintained the geometric structure of the target domain. Lu et al.\cite{111} introduced a multi-source compensation network (MSCN) that integrates a cross-domain alignment module with a classifier compensation mechanism to mitigate domain offset and enhance category-level semantic alignment. Similarly, an attention-driven multi-scale residual adaptation network (AMRAN) was proposed by Zhu et al.~\cite{112}, which includes an edge distribution alignment module, an attention module, and a multi-scale adaptation module. Geng et al.~\cite{113} proposed a deep joint distribution adaptation network (DJDAN), which simultaneously aligns the marginal and conditional distributions for transfer learning in SAR image classification.

In recent years, self-supervised learning (SSL) has gradually become an effective strategy when labeled data is lacking~\cite{102}, showing great potential for feature representation learning. Zhao et al.~\cite{114} proposed a contrastive learning and domain adaptation layer (CLA) framework for leaf disease identification. This method combines self-supervised pre-training and domain adaptive fine-tuning, and is divided into two stages: the first stage trains the encoder on a large number of unlabeled images thru contrastive learning; the second stage uses the domain adaptive layer to jointly optimize the supervised loss and the domain alignment loss on a small amount of labeled data. The total loss function is: \begin{equation} \label{eq:dal_loss} L_{\text{DAL}} = L_{\text{supervised}} + \lambda \cdot L_{\text{MMD}} \end{equation} where $L_{\text{supervised}}$ represents the supervised classification loss, $L_{\text{MMD}}$ represents the alignment loss based on MMD, and $\lambda$ is the weighting coefficient.

The experimental results show that CLA is significantly better than the comparison model in domain alignment and classification accuracy, and the highest accuracy rate is 90.52\%. Ablation experiments and sensitivity analysis identified the key factors behind the performance improvement. The framework combines self-supervised contrastive learning with domain adaptation. It provides an efficient and scalable solution for leaf disease identification under low-annotation conditions.


\subsection{Summary of Deep Domain Adaptation (DA) Methods}

With the rapid development of deep learning, domain adaptation (DA) methods based on deep neural networks (DNN) have shown great potential in agricultural image analysis. Thru end-to-end feature extraction and distribution alignment, these methods can automatically learn domain-invariant features, effectively alleviating the domain shift caused by sensor differences, environmental complexity, and different crop growth stages. In general, Deep DA methods are generally classified into three types: \textit{supervised}, \textit{semi-supervised}, and \textit{unsupervised}.

In \textbf{supervised} DA, the model relies on labeled samples from the source and target domains, and achieves robust transfer by jointly optimizing the classification loss and the domain alignment loss. On this basis, \textbf{semi-supervised} DA incorporates a limited set of labeled samples from the target domain, taking into account the labeling cost and performance, and is especially suitable for agricultural labeling scarce scenarios.

\textbf{Unsupervised} DA does not require target domain labels. It usually uses feature alignment or generative models (such as GAN-based style transfer) to achieve cross-domain adaptation, which is suitable for large-scale agricultural tasks that are difficult to label.

The core advantage of deep DA is that it can process high-dimensional, multi-modal agricultural data and maintain task-related semantics. However, it still faces problems such as large amount of calculation, long training time, and insufficient interpretability, and its deployment is limited in resource-limited or real-time applications.

Future research directions include lightweight networks, interpretable learning mechanisms, and hardware acceleration to further enhance the application value of deep DA in agriculture. With the continuous improvement of the method, these technologies are expected to become important support tools for smart agriculture and precision agriculture, covering a variety of tasks from crop monitoring to disease diagnosis.

\section{Experimental Results: Evaluation, Comparison and Analysis}

To provide a comprehensive evaluation of the impact of deep DA methods in agriculture, this section conducts experiments on three representative tasks: \textit{plant disease detection}, \textit{crop yield prediction}, and \textit{remote sensing agricultural plot extraction}. The results include both quantitative indicators and qualitative visualization comparisons. To ensure fairness, all methods are evaluated on a common benchmark dataset, and the best results are marked in \textbf{bold}.

\subsection{Dataset Overview}

The experiments use a diverse set of benchmark datasets to ensure the comprehensiveness and robustness of the results.

In terms of \textbf{remote sensing agricultural mapping}: 
\begin{itemize} 
\item \textbf{DeepGlobe}: Used for urban and rural land cover classification and segmentation, the data comes from DeepGlobe 2018 Challenge.
\item \textbf{GID}: A large-scale high-resolution land use/cover mapping dataset, the download link is \url{https://captain-whu.github.io/GID/}.
\item \textbf{LoveDA}: For semantic segmentation tasks, covering agriculture and geographic information extraction, it is open to \url{https://github.com/Junjue-Wang/LoveDA}.
\end{itemize}

In terms of \textbf{plant disease detection and classification}: 
\begin{itemize} 
\item \textbf{PlantVillage}: It is often used for disease identification in laboratory environments,and the dataset can be accessed publicly on Kaggle at \url{https://www.kaggle.com/datasets/emmarex/plantdisease} .
\item \textbf{PlantDoc}: Covers real-field disease diagnosis, available from the PlantDoc-Dataset GitHub repository.
\item \textbf{Plant-Pathology}: From the Kaggle FGVC8 competition, it is used for disease identification and severity estimation.
\item \textbf{Corn-Leaf-Diseases}: A corn disease classification dataset, publicly available on Kaggle.
\item \textbf{Tomato-Leaf-Diseases}: Tomato disease monitoring dataset, available on Kaggle and Mendeley Data.
\end{itemize}
For \textbf{agricultural yield prediction}: \begin{itemize} \item \textbf{NASS}: Tabular data released by the USDA NASS for yield prediction and cross-domain productivity modeling (\url{https://www.nass.usda.gov}).
\end{itemize}

Table~\ref{tab:datasets} summarizes the spatial resolution, sensor type, image size, and division strategy of each dataset. They cover a wide range of agricultural scenes, sensor modes and domain shift situations, providing a representative performance evaluation benchmark for DA methods.

In the performance evaluation of different agricultural tasks, this paper adopts a variety of indicators for classification, regression and semantic segmentation, and the specific definitions and mathematical forms will be given below.


\begin{table*}[ht]
\centering
\caption{Summary of datasets used in the experiments. ``N/A'' indicates information not explicitly defined in the dataset description.}
\label{tab:datasets}
\renewcommand{\arraystretch}{1.25}
\setlength{\tabcolsep}{6pt}
\resizebox{\textwidth}{!}{
\begin{tabular}{lcccccc}
\toprule
\textbf{Dataset} & \textbf{Resolution} & \textbf{Sensor} & \textbf{Image Size} & \textbf{Training Set} & \textbf{Test Set} \\
\midrule
DeepGlobe            & 0.5 m/pixel         & WorldView-2       & 2448 $\times$ 2448    & 30,470        & N/A           \\
GID                  & 4 m/pixel           & GF-2              & 6800 $\times$ 7200    & 73,490        & 734           \\
LoveDA               & 0.3 m/pixel         & Spaceborne        & 1024 $\times$ 1024    & 15,829        & 324           \\
PlantVillage         & 256 $\times$ 256    & RGB               & Fixed size            & 50,000        & 5,000         \\
PlantDoc             & Variable            & RGB               & Variable              & 2,000         & 500           \\
Plant-Pathology      & 2048 $\times$ 1365  & RGB               & Fixed size            & 2,921         & 730           \\
Corn-Leaf-Diseases   & 256 $\times$ 256    & RGB               & Fixed size            & 2,500         & 500           \\
Tomato-Leaf-Diseases & 256 $\times$ 256    & RGB               & Fixed size            & 4,000         & 1,000         \\
NASS                 & N/A                 & Tabular Data      & Multi-year (regionwise) & N/A         & N/A           \\
\bottomrule
\end{tabular}
}
\end{table*}

\subsection{Evaluation Metrics}

\paragraph{Average Accuracy (AA)} \begin{equation} \text{AA} = \frac{1}{C} \sum_{i=1}^{C} \frac{N_{ii}}{\sum_{j=1}^{C} N_{ij}} \end{equation} \textit{where:} $C$ is the total number of categories, and $N_{ij}$ represents the number of samples whose true label is $i$ but predicted as $j$. AA assigns the same weight to each category, which alleviates the problem of category imbalance in agricultural multi-classification data sets.

\paragraph{Root Mean Square Error (RMSE)} \begin{equation} \text{RMSE} = \sqrt{ \frac{1}{n} \sum_{i=1}^{n} (y_i - \hat{y}_i)^2 } \end{equation} \textit{where:} $y_i$ is the true value and $\hat{y}_i$ is the predicted value. RMSE has the same unit as the target variable, and is a commonly used indicator for regression tasks such as yield prediction.

\paragraph{Coefficient of determination ($R^2$)} \begin{equation} R^2 = 1 - \frac{\sum_{i=1}^{n} (y_i - \hat{y}i)^2}{\sum{i=1}^{n} (y_i - \bar{y})^2} \end{equation} \textit{where:} $\bar{y}$ is the mean of the true values. $R^2$ represents the proportion of the variance of the independent variable explained by the dependent variable, and is an intuitive indicator for measuring the goodness of fit of the model.

\paragraph{Intersection over Union (IoU)} \begin{equation} \text{IoU} = \frac{|P \cap G|}{|P \cup G|} \end{equation} \textit{where:} $P$ is the predicted region and $G$ is the ground truth region. IoU is often used for semantic segmentation evaluation, especially suitable for farmland feature mapping.
\paragraph{Completeness (COM)} \begin{equation} \text{COM} = \frac{\text{TP}}{\text{TP} + \text{FN}} \end{equation} COM measures the proportion of true farmland pixels that are correctly identified.

COR represents the proportion of pixels that are truly correct among the pixels predicted to be farmland, and is used to evaluate false detection.

\paragraph{Segmentation F1 score} \begin{equation} \text{F1}_{\text{seg}} = \frac{2 \cdot \text{COM} \cdot \text{COR}}{\text{COM} + \text{COR}} \end{equation} This indicator provides a balanced evaluation of the quality of agricultural image segmentation by combining completeness and correctness.

\subsection{Plant Disease Detection Experiment}

\begin{figure}[ht] \centering \includegraphics[width=1\linewidth]{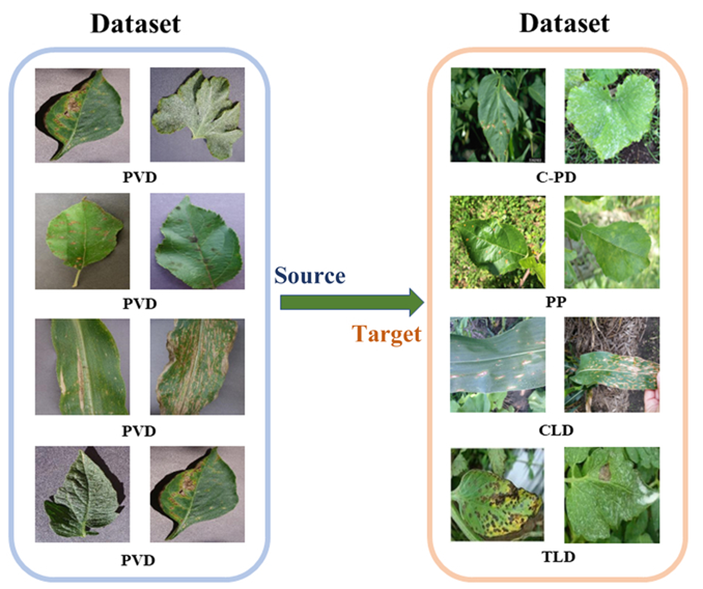} \caption{Domain adaptation configurations for the four groups of experiments. Left: source domain dataset (PVD); right: target domain dataset.} \label{fig15} \end{figure}

In the plant disease detection task, this paper designed four groups of experiments based on five benchmark datasets PlantVillage (PVD)\cite{115}, PlantDoc\cite{116}, Plant-Pathology~\cite{117}, Corn-Leaf-Diseases and Tomato-Leaf-Diseases (see Figure~\ref{fig15}). Among them, PlantVillage is collected under controlled conditions and has a clean background. It is uniformly set as the source domain to ensure the stability of model training.

The plant disease detection process includes four steps: (1) obtaining images from different domains; (2) performing preprocessing, such as scaling, normalization, and enhancement, to reduce differences in resolution and illumination; (3) extracting features using convolutional neural networks; and (4) completing classification thru domain adaptive models. The experiment adopts the MSUN (Multi-Representation Subdomain Adaptation Network with Uncertainty Regularization) method. The model contains three key modules: multi-representation feature extraction, subdomain alignment, and uncertainty regularization. The former learns both global and fine-grained features, the middle one reduces the inter-class distribution difference, and the latter improves the stability of low-confidence sample prediction. MSUN is particularly suitable for agricultural data, where disease symptoms may be similar between different diseases, and may also change greatly with the growth stage.

Table~\ref{tab:msun-results} shows the classification accuracy of eight models (Baseline, DAAN, D-CORAL, DAN, DANN, MRAN, DSAN and MSUN) on four transfer tasks (C-PD, PVD-PP, PVD-CLD, PVD-TLD). The results show that the single-crop transfer tasks (such as PVD-CLD and PVD-TLD) are easier to obtain higher accuracy than the multi-crop mixed transfer task (C-PD) because the intra-class difference is small. MSUN performs best in all scenarios, and its advantages mainly come from the multi-representation learning and sub-domain alignment strategy, which effectively alleviates the problems of inter-class similarity and intra-class difference.

In the cross-domain disease detection experiment, MSUN has a significant improvement in all configurations. In the C-PD task, the average accuracy rate is 56.06\%, which is 25 percentage points higher than the baseline, and is better than DSAN and MRAN. In the PVD-PP task, MSUN achieved the best results in all three types of diseases, with an overall accuracy rate of 72.31\%, exceeding MRAN's 67.45\% and DANN's 50.04\%.
In the PVD-CLD task, its average accuracy was 96.78\%, of which "Blight" was 93.81\% and "Healthy" was 84.77\%. In the more challenging PVD-TLD task, MSUN still achieved 50.58\% even with fine-grained categories and unbalanced samples, in which the "Septoria spot" class (low sample) was 80.79\%.

In summary, MSUN significantly improves the classification accuracy, robustness and generalization ability in the cross-domain detection of plant diseases, and can be better than the traditional methods under the conditions of domain offset, category imbalance and complex inter-class/intra-class.
\begin{table*}[ht]
\centering
\caption{Classification accuracy (\%) across four experimental setups: C-PD, PVD-PP, PVD-CLD, and PVD-TLD. The highest accuracy for each class/task is highlighted in \textbf{bold}.}
\label{tab:msun-results}
\renewcommand{\arraystretch}{1.2}
\setlength{\tabcolsep}{5pt}
\resizebox{\textwidth}{!}{
\begin{tabular}{llcccccccc}
\toprule
\textbf{Task} & \textbf{Class / Crop} & \textbf{Baseline} & \textbf{DAAN} & \textbf{D-CORAL} & \textbf{DAN} & \textbf{DANN} & \textbf{MRAN} & \textbf{DSAN} & \textbf{MSUN} \\
\midrule
\multirow{7}{*}{C-PD}      
& Apple         & 67.38 & 68.64 & 69.06 & 70.29 & 69.63 & 68.98 & 65.51 & \textbf{71.45} \\
& Corn          & 50.07 & 48.81 & 49.11 & 52.15 & 48.11 & 49.87 & 51.39 & \textbf{56.46} \\
& Grape         & 71.36 & 63.93 & 79.79 & 78.24 & 83.41 & 82.38 & 79.79 & \textbf{85.49} \\
& Pepper bell   & 84.19 & 87.79 & 86.75 & 90.91 & 90.90 & 92.46 & 90.91 & \textbf{93.49} \\
& Potato        & 50.34 & 63.66 & 66.06 & 64.72 & 62.33 & 57.82 & 64.19 & \textbf{67.42} \\
& Tomato        & 30.17 & 39.82 & 43.32 & 46.23 & 49.51 & 49.34 & 45.75 & \textbf{48.67} \\
& \textbf{Average} & 30.78 & 33.08 & 33.37 & 33.79 & 39.71 & 39.84 & 42.87 & \textbf{56.06} \\
\midrule
\multirow{4}{*}{PVD-PP}    
& Gray spot     & 79.54 & 66.16 & 77.56 & \textbf{89.02} & 85.12 & 81.28 & 83.96 & 87.32 \\
& Rust          & 24.10 & 55.81 & 33.42 & 34.99 & 30.26 & 52.57 & 35.42 & \textbf{57.71} \\
& Healthy       & 11.02 & 36.64 & 48.05 & 11.84 & 28.71 & 70.87 & 51.24 & \textbf{73.86} \\
& \textbf{Average} & 39.27 & 53.75 & 52.41 & 48.89 & 50.04 & 67.45 & 56.41 & \textbf{72.31} \\
\midrule
\multirow{5}{*}{PVD-CLD}   
& Gray spot     & 61.49 & 78.57 & 79.44 & 68.29 & 76.66 & 79.22 & \textbf{81.71} & 80.09 \\
& Blight        & 86.75 & 92.27 & 91.88 & 91.50 & 91.73 & 92.05 & 91.58 & \textbf{93.81} \\
& Healthy       & 79.05 & 84.29 & 82.64 & 81.33 & 82.64 & 83.61 & 83.42 & \textbf{84.77} \\
& Rust          & 99.18 & \textbf{100.0} & \textbf{100.0} & \textbf{100.0} & \textbf{100.0} & \textbf{100.0} & \textbf{100.0} & \textbf{100.0} \\
& \textbf{Average} & 75.96 & 90.35 & 89.90 & 87.89 & 89.47 & 91.06 & 90.69 & \textbf{96.78} \\
\midrule
\multirow{6}{*}{PVD-TLD}   
& Bacterial spot & 5.34  & 10.23 & 12.50 & 12.22 & 8.24  & 18.47 & 10.51 & \textbf{29.27} \\
& Healthy        & 70.61 & 83.75 & 85.81 & 83.41 & \textbf{86.92} & 75.92 & 92.03 & 78.67 \\
& Late blight    & 79.05 & 3.97  & 1.64  & 0.93  & 3.50  & 2.11  & 1.64  & \textbf{3.28} \\
& Mold leaf      & 25.03 & 33.15 & \textbf{37.43} & 38.18 & 38.73 & 33.15 & 29.61 & 32.77 \\
& Septoria spot  & 6.36  & 5.16  & 18.73 & 67.11 & 4.13  & 78.64 & 0.29  & \textbf{80.79} \\
& \textbf{Average} & 21.46 & 30.96 & 32.31 & 42.80 & 28.71 & 44.86 & 27.57 & \textbf{50.58} \\
\bottomrule
\end{tabular}
}
\end{table*}

\setlength{\tabcolsep}{8pt}
\renewcommand{\arraystretch}{1.2}
\begin{table*}[ht]
\centering
\caption{Performance comparison in crop yield prediction across years and transfer settings. Best results are in \textbf{bold}.}
\label{tab:yield}
\begin{tabular}{cccccccc}
\toprule
\textbf{Year} & \textbf{Transfer} & \textbf{Metric} & \textbf{RF} & \textbf{DNN} & \textbf{DANN} & \textbf{ADANN} \\
\midrule
\multirow{8}{*}{2016} 
& GP$\rightarrow$GP   & $R^2$   & 0.75 & 0.81 & 0.64 & \textbf{0.85} \\
&                     & RMSE    & 1.39 & 1.18 & 1.65 & \textbf{1.05} \\
& ETF$\rightarrow$ETF & $R^2$   & 0.51 & 0.55 & 0.47 & \textbf{0.62} \\
&                     & RMSE    & 1.13 & 1.01 & 1.18 & \textbf{0.92} \\
& GP$\rightarrow$ETF  & $R^2$   & 0.54 & 0.43 & 0.44 & \textbf{0.67} \\
&                     & RMSE    & 1.02 & 1.22 & 1.21 & \textbf{0.86} \\
& ETF$\rightarrow$GP  & $R^2$   & 0.48 & 0.55 & 0.68 & \textbf{0.76} \\
&                     & RMSE    & 1.98 & 1.82 & 1.54 & \textbf{1.35} \\
\midrule
\multirow{8}{*}{2017} 
& GP$\rightarrow$GP   & $R^2$   & 0.73 & 0.75 & 0.74 & \textbf{0.77} \\
&                     & RMSE    & 1.34 & 1.26 & 1.30 & \textbf{1.24} \\
& ETF$\rightarrow$ETF & $R^2$   & 0.61 & 0.75 & 0.64 & \textbf{0.78} \\
&                     & RMSE    & 1.23 & 0.97 & 1.17 & \textbf{0.91} \\
& GP$\rightarrow$ETF  & $R^2$   & 0.58 & 0.52 & 0.35 & \textbf{0.73} \\
&                     & RMSE    & 1.26 & 1.36 & 1.58 & \textbf{1.03} \\
& ETF$\rightarrow$GP  & $R^2$   & 0.52 & 0.64 & 0.68 & \textbf{0.77} \\
&                     & RMSE    & 1.78 & 1.54 & 1.45 & \textbf{1.23} \\
\midrule
\multirow{8}{*}{2018} 
& GP$\rightarrow$GP   & $R^2$   & 0.76 & 0.79 & 0.81 & \textbf{0.84} \\
&                     & RMSE    & 1.11 & 1.05 & 1.00 & \textbf{0.91} \\
& ETF$\rightarrow$ETF & $R^2$   & 0.54 & 0.56 & 0.51 & \textbf{0.65} \\
&                     & RMSE    & 0.97 & 0.95 & 1.00 & \textbf{0.84} \\
& GP$\rightarrow$ETF  & $R^2$   & 0.49 & 0.30 & 0.33 & \textbf{0.57} \\
&                     & RMSE    & 1.01 & 1.20 & 1.17 & \textbf{0.97} \\
& ETF$\rightarrow$GP  & $R^2$   & 0.53 & 0.49 & 0.75 & \textbf{0.78} \\
&                     & RMSE    & 1.57 & 1.63 & 1.14 & \textbf{1.07} \\
\midrule
\multirow{8}{*}{2019} 
& GP$\rightarrow$GP   & $R^2$   & 0.74 & 0.73 & 0.53 & \textbf{0.76} \\
&                     & RMSE    & 1.14 & 1.15 & 1.53 & \textbf{1.08} \\
& ETF$\rightarrow$ETF & $R^2$   & 0.63 & 0.71 & 0.64 & \textbf{0.66} \\
&                     & RMSE    & 1.09 & 0.96 & 1.08 & \textbf{1.05} \\
& GP$\rightarrow$ETF  & $R^2$   & 0.52 & 0.56 & 0.62 & \textbf{0.68} \\
&                     & RMSE    & 1.25 & 1.19 & 1.11 & \textbf{1.02} \\
& ETF$\rightarrow$GP  & $R^2$   & 0.19 & 0.34 & 0.56 & \textbf{0.73} \\
&                     & RMSE    & 2.00 & 1.80 & 1.47 & \textbf{1.15} \\
\bottomrule
\end{tabular}
\end{table*}

\subsection{Crop Yield Prediction Experiments}

\begin{figure}[ht] \centering \includegraphics[width=1\linewidth]{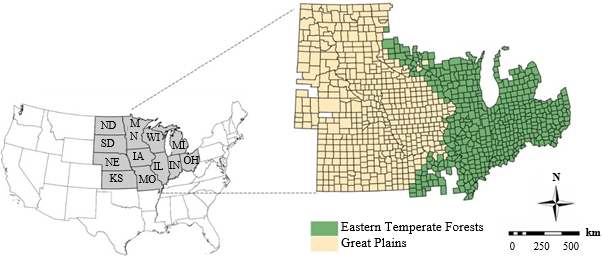} \caption{Two ecoregions used in the cross-domain yield prediction experiment: Eastern Temperate Forest (ETF) and Great Plains (GP). Figure from Ma \textit{et al.}~\cite{143}.}   \label{fig16}
\end{figure}

To evaluate the cross-domain adaptability in the agricultural regression task, this study selected the US Corn Belt, the world's largest corn production area, which has a wealth of historical yield records. The model training combines remote sensing products, meteorological observations, and crop yield statistics, all of which come from the National Agricultural Statistics Service (NASS, 2020) of the U.S. Department of Agriculture, as shown in Figure~\ref{fig16}.

The ADANN (Adaptive Domain-Adversarial Neural Network) framework used in this paper introduces a dynamic weighted loss function on the basis of DANN to balance the yield prediction target and the domain alignment target. This mechanism can better adapt to the distribution differences of different ecological zones when the labeled data is limited. Its structure consists of a shared feature extraction layer and an adversarial training domain discriminator, combined with adaptive weights to ensure prediction accuracy~\cite{143}.

The experimental design includes two types of scenarios: \textit{local} (within the region) and \textit{cross-domain} (between regions) prediction. The four models compared are: Random Forest (RF), Deep Neural Network (DNN), Domain Adversarial Neural Network (DANN) and the improved version ADANN. The experimental areas include the Eastern Temperate Forest (ETF) and the Great Plains (GP).

Table~\ref{tab:yield} shows the results of local and cross-domain prediction in four years. In local prediction (e.g. GP$\rightarrow$GP, ETF$\rightarrow$ETF), DNN and ADANN perform better than RF and DANN. In cross-domain prediction, ADANN always obtains higher $R^2$ and lower RMSE, showing stronger domain-invariant feature learning ability.

For example, in the 2019 ETF$\rightarrow$GP task, the $R^2$ of RF was only 0.19, while ADANN reached 0.73, RMSE=1.15, which was significantly better than DANN (0.56, 1.47). DNN and RF are stable in local prediction, but their accuracy decreases significantly when domain migration occurs. After introducing adversarial training, DANN's robustness is improved, while ADANN performs best and consistently.

The abnormal climate also affected the model performance. In 2019, the abnormal wetness in some areas of the corn belt led to a decline in the correlation between the spectral index and the yield, which aggravated the domain difference. However, ADANN still maintained stable prediction, showing its adaptability to the environmental fluctuations across years.

\begin{figure}[ht] \centering \includegraphics[width=1\linewidth]{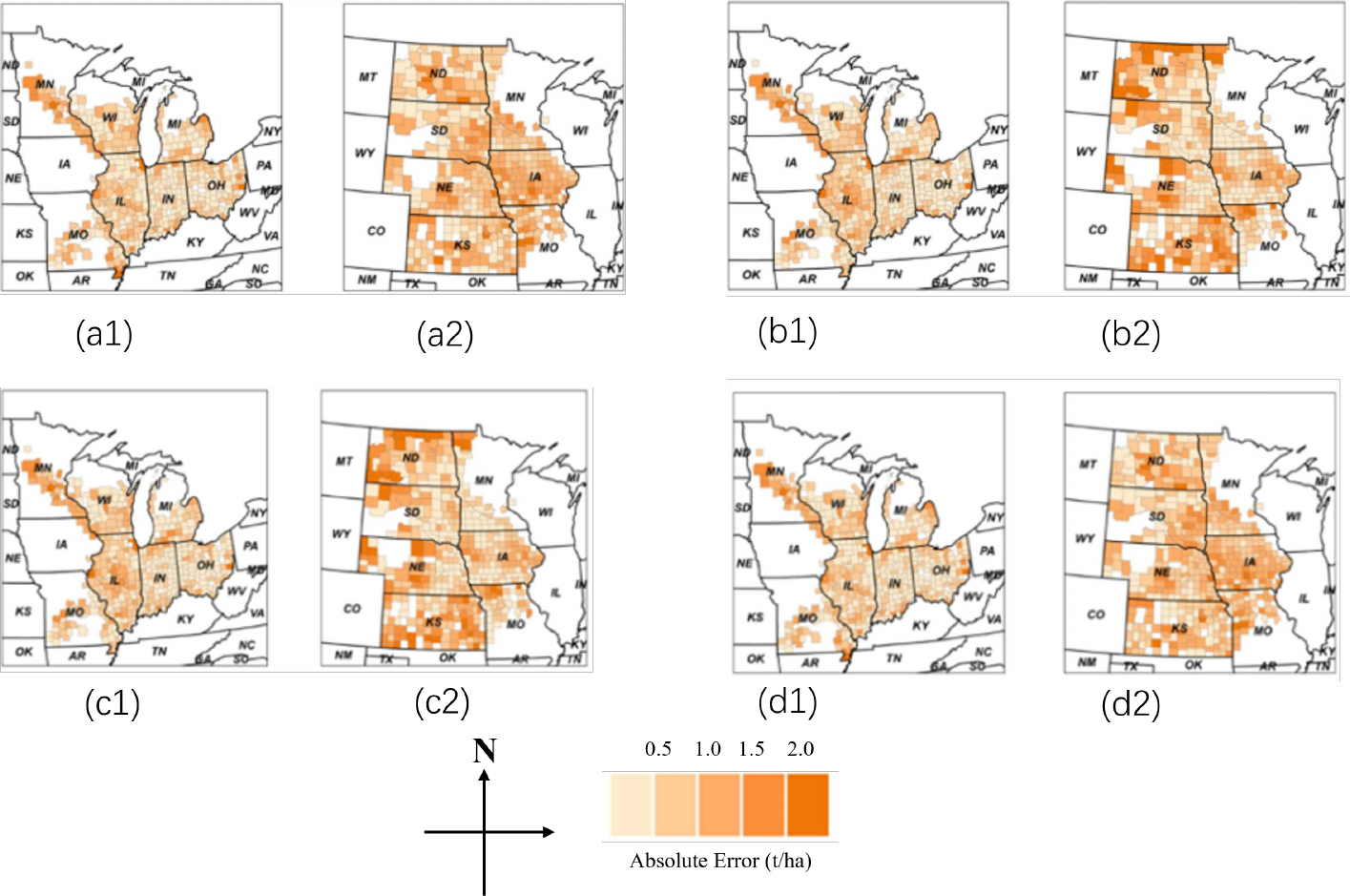} \caption{The distribution of mean absolute error of GP$\rightarrow$ETF and ETF$\rightarrow$GP from 2016 to 2019. (a1, a2) RF, (b1, b2) DNN, (c1, c2) DANN, (d1, d2) ADANN. Figure from Ma \textit{et al.}~\cite{143}.}\label{fig17}
\end{figure}

Figure~\ref{fig17} shows the distribution of mean absolute error for GP$\rightarrow$ETF and ETF$\rightarrow$GP. Traditional models show high error clustering in states far from the source domain, while adversarial methods (especially ADANN) significantly reduce spatial bias in major agricultural areas.

In summary, ADANN is superior to other models in both local and cross-domain prediction, which proves its applicability under distribution differences. By balancing adversarial feature alignment and prediction accuracy, this method provides a strong reference for cross-regional agricultural yield prediction.


\subsection{Remote Sensing-Based Agricultural Land Extraction Experiments}

\begin{figure}[ht] \centering \includegraphics[width=1\linewidth]{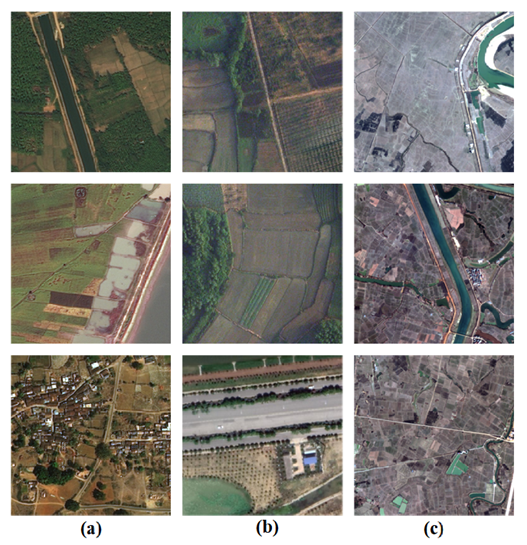} \caption{Examples of three agricultural land segmentation datasets: (a) DeepGlobe (0.5 m/pixel); (b) LoveDA (0.3 m/pixel); (c) GID (4 m/pixel).}    \label{fig18}
\end{figure}

To evaluate the domain adaptation performance of agricultural land segmentation in remote sensing images, this paper selects six typical methods: baseline \textit{source-only} model (no adaptation training), AdaptSegNet~\cite{83}, ADVENT~\cite{84}, BDL~\cite{85}, IntraDA~\cite{52} and TransFusion-DualDA~\cite{65}. The experiments are carried out on three benchmark datasets, DeepGlobe, LoveDA and GID, which differ significantly in resolution and scene complexity (see Figure~\ref{fig18}).

Before giving the results, the process of TransFusion-DualDA is introduced first. The framework adopts the Transformer-based segmentation backbone, taking into account local details and global context. In the training, the features of the source domain and the target domain are aligned thru adversarial learning or self-training, so as to ensure the accuracy of cross-resolution and scene migration. Its multi-scale feature fusion and context modeling further reduce the domain difference, especially suitable for DeepGlobe, LoveDA and GID and other data sets with obvious differences.

Table~\ref{tab:uda-results} shows the quantitative results of three types of transfer tasks: DeepGlobe$\rightarrow$LoveDA, GID$\rightarrow$LoveDA and DeepGlobe$\rightarrow$GID.

TransFusion-DualDA achieves the highest IoU, COM and F1 values in all scenarios.
In the DeepGlobe$\rightarrow$LoveDA task, the IoU reaches 55.76\% and the F1 reaches 67.75\%, which is about 5–10 percentage points higher than BDL and IntraDA. This is due to the advantages of the Transformer structure in global modeling and multi-scale feature fusion.

In the GID$\rightarrow$LoveDA migration, the COM of TransFusion-DualDA reached 92.11\%, which was more than 32\% higher than the source-only baseline. This indicates that it can effectively capture boundary details in the migration from low resolution (4 m/pixel) to ultrahigh resolution (0.3 m/pixel), while traditional methods often perform poorly due to scale mismatch.

In the DeepGlobe$\rightarrow$GID (high-resolution to low-resolution) task, the model achieved F1=62.44\% and IoU=49.55\%. Although the increase was smaller than that of IntraDA and BDL, the results reflected the problem of missing details caused by the low resolution of GID, which limited the transfer effect of high-resolution features.

\begin{figure}[ht] \centering \includegraphics[width=1\linewidth]{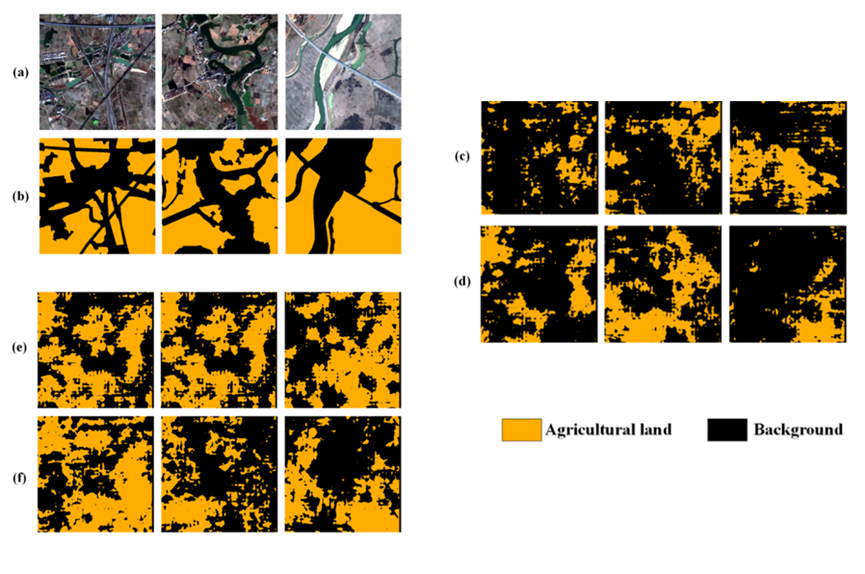} \caption{Segmentation results of DeepGlobe$\rightarrow$GID. (a) Target image; (b) Ground truth; (c) Source-only; (d) ADVENT; (e) IntraDA; (f) TransFusion-DualDA.}   \label{fig19}
\end{figure}

Figure~\ref{fig19} shows the qualitative results. The source-only model (c) is seriously confused between farmland and background, and the structural loss is obvious. ADVENT (d) has been improved thru adversarial learning, and IntraDA (e) has optimized the boundary by using intra-class consistency, but there are still fuzzy and fragmented phenomena.

In contrast, the segmentation map generated by TransFusion-DualDA(f) is the closest to the ground truth, and the overall structure and details are preserved. Its advantage comes from the effective long-distance dependency modeling and full-scale feature fusion. Although there are still slight artifacts on the edges, the overall performance is better than the comparison method.

In summary, TransFusion-DualDA shows good generalization ability under different resolutions and land types. Its stable advantage in shape-sensitive indicators such as IoU and COM verifies its robustness in boundary maintenance, and it is a powerful solution for unsupervised domain adaptation of agricultural land segmentation in the field of remote sensing.


\setlength{\tabcolsep}{8pt}
\renewcommand{\arraystretch}{1.2}
\begin{table*}[ht]\centering
    \caption{Comparison of domain adaptation methods for agricultural land extraction under different dataset transfer settings.Best results are in \textbf{bold}.}\label{tab:uda-results}
\begin{tabular}{lccccc}
\toprule
\textbf{Source $\rightarrow$ Target} & \textbf{Method} & \textbf{IoU} & \textbf{COM} & \textbf{COR} & \textbf{F1} \\ 
\midrule
\multirow{6}{*}{DeepGlobe $\rightarrow$ LoveDA} 
& Source-only & 36.33 & 45.90 & 62.86 & 49.11 \\
& AdaptSegNet & 46.92 & 74.48 & 55.19 & 60.40 \\
& ADVENT      & 49.39 & 74.62 & 58.09 & 61.17 \\
& BDL         & 52.23 & 79.75 & 59.01 & 65.33 \\
& IntraDA     & 51.71 & 82.86 & 57.34 & 64.59 \\
& TransFusion-DualDA & \textbf{55.76} & \textbf{81.37} & \textbf{62.55} & \textbf{67.75} \\
\midrule
\multirow{6}{*}{GID $\rightarrow$ LoveDA} 
& Source-only & 36.23 & 40.14 & \textbf{83.02} & 46.03 \\
& AdaptSegNet & 42.93 & 54.22 & 71.30 & 55.45 \\
& ADVENT      & 45.04 & 60.76 & 65.02 & 58.25 \\
& BDL         & 44.59 & 58.23 & 67.45 & 57.73 \\
& IntraDA     & 48.25 & 60.37 & 72.90 & 60.80 \\
& TransFusion-DualDA & \textbf{53.47} & \textbf{92.11} & 56.89 & \textbf{66.04} \\
\midrule
\multirow{6}{*}{DeepGlobe $\rightarrow$ GID} 
& Source-only & 26.99 & 39.20 & 62.88 & 36.60 \\
& AdaptSegNet & 43.10 & 68.61 & 59.41 & 56.07 \\
& ADVENT      & 46.00 & 77.44 & 57.40 & 58.94 \\
& BDL         & 46.35 & 75.79 & 58.37 & 59.51 \\
& IntraDA     & 47.63 & 68.49 & \textbf{60.29} & 61.47 \\
& TransFusion-DualDA & \textbf{49.55} & \textbf{88.55} & 54.54 & \textbf{62.44} \\
\bottomrule
\end{tabular}
\end{table*}

\subsection{Summary of Experimental Results}

In the cross-domain disease detection experiment, MSUN showed a significant improvement in all tasks. In the multi-crop mixed migration task C-PD, the average accuracy rate reached 56.06\%, which was 25 percentage points higher than the baseline, and better than DSAN and MRAN. In the PVD-PP task, MSUN achieved the best results in all three types of diseases, with an overall accuracy of 72.31\%, exceeding MRAN (67.45\%) and DANN (50.04\%).

The advantages of MSUN come from its structural design. The multi-representation module can learn global and fine-grained features to reduce inter-domain and inter-class differences. The sub-domain alignment mechanism ensures semantic fine-grained consistency. The added uncertainty regularization can suppress pseudo-label noise. These designs enable MSUN to take the lead in four public benchmark datasets, showing strong generalization and accuracy in multi-crop disease diagnosis.

In the cross-ecoregion yield prediction experiment in the US Corn Belt, four models, RF, DNN, DANN, and ADANN, were evaluated. ADANN performed best in both in-domain and cross-domain predictions from 2016 to 2019. For example, in the 2019 in-domain task GP$\rightarrow$GP, the $R^2$ of RF and DNN were 0.76 and 0.79, respectively, while ADANN reached 0.85. In the cross-domain task GP$\rightarrow$ETF, the performance of RF and DNN decreased significantly (0.49 and 0.30), while ADANN maintained a robust $R^2=0.63$, which was significantly better than DANN (0.58).

The interpretability analysis shows that ADANN reduces the domain shift by balancing the prediction loss and the alignment loss. The spatial error map shows that RF and DNN have obvious errors in Michigan, Ohio and other places. DANN reduces these errors thru adversarial training. ADANN further reduces the error in key areas such as Illinois, showing stronger cross-regional robustness.

In the land extraction experiment, TransFusion-DualDA outperforms other methods. In the DeepGlobe$\rightarrow$LoveDA migration, the IoU is 55.763\% and the F1 is 67.750\%, which is 6–7 percentage points higher than ADVENT and AdaptSegNet. In the more difficult low-resolution to high-resolution migration (GID$\rightarrow$LoveDA), the IoU is increased by 5.2\% and the F1 is increased by 6\%. This shows that the Transformer backbone and multi-scale fusion help to capture contextual and structural information.

In the reverse transfer DeepGlobe$\rightarrow$GID (high to low resolution), TransFusion-DualDA also performed well (F1 = 62.439\%), which is better than BDL and other methods. The visual results show that its boundary and region accuracy are better than ADVENT and IntraDA, but there is still noise in the target domain. Local feature alignment is still challenging at low resolution.

In general, TransFusion-DualDA combines adversarial training with entropy-based subdomain division, which effectively reduces the domain difference in complex situations such as urban-rural transition zones. Future research can further optimize the global feature extraction of high-resolution images to enhance the adaptability in dynamic remote sensing.

\section{Challenges and Future Research Directions}

Although domain adaptation in agricultural intelligence has made some progress, there are still difficulties in data acquisition, model design and practical implementation. Agricultural data often presents obvious spatiotemporal heterogeneity and multi-modal characteristics, and the number of annotations is limited~\cite{kamilaris2018deep,zhang2021review}. Domain adaptation can alleviate the domain differences caused by environmental changes and insufficient annotation to a certain extent~\cite{tzeng2017adversarial,ganin2016domain}. However, there are still problems such as large computational overhead, limited interpretability, and insufficient robustness in dynamic environments.

\subsection{Adaptive Impairment in Field Scenes}

The following problems exist in the actual agricultural deployment of deep learning:

\textbf{1) Multi-scale spatio-temporal data fusion:} Agricultural monitoring combines satellite images (10–100 m/pixel), UAV images (0.1–1 m/pixel) and ground images (centimeter-level resolution), and the acquisition time is different~\cite{yuan2020deep,zhong2019deep}. Existing 3D convolution and Transformer models have difficulty aligning features on such a wide spatiotemporal scale. The difference in resolution and revisit frequency reduces consistency, and the model is difficult to expand from the cell to the regional or national scale.

\textbf{2) Insufficient high-quality labeled data:} In many cases, there are less than 50 labeled samples per class, especially for rare diseases or stress types~\cite{lu2021survey}. In low-resource scenarios, the performance of traditional fine-tuning or transfer learning decreases by more than 30\%. Data collection is limited by seasons and the cost of expert annotation.

\textbf{3) Limited multi-modal fusion:} Hyperspectral, multispectral, LiDAR and thermal imaging can provide complementary information, but field integration is difficult. Illumination changes (e.g., clouds causing ±30\% fluctuations) introduce spectral distortion~\cite{pu2021hyperspectral}. The spatial error between the air and ground sensors (often exceeding 0.5 m) limits the fusion accuracy~\cite{ma2019deep}.

\textbf{4) Model interpretability and credibility:} Deep models are often regarded as "black boxes"~\cite{samek2017explainable}. The lack of correspondence between predictions and plant traits or physical processes affects decision-making applications. Robustness under out-of-distribution or adversarial conditions is rarely assessed, which makes the reliability of yield prediction or disease early warning questionable.

\subsection{Future Research Directions}

Solving the above problems requires the efforts of all parties:

\textbf{(1) Lightweight spatio-temporal modeling:} Develop a compact model with parameters less than 1M. Combine crop growth and canopy reflection model~\cite{jacquemoud1990prospect}. Using separable convolution and time decomposition~\cite{wang2021spatio}, the energy consumption can be reduced to 20\% of the current 3D CNN, supporting the deployment of UAVs and edge devices.

\textbf{(2) Weakly supervised and self-supervised learning:} Pseudo-labeling strategies are designed based on phenological stages to reduce the dependence on large-scale labeling. Curriculum learning based on crop growth stages gradually increases the complexity, and the accuracy is still maintained at >85\% when the number of labeled samples is less than 100 ~\cite{zoph2020rethinking}. Cross-season consistency and joint training can stabilize performance.

\textbf{(3) Multi-modal fusion and physical consistency:} Spectral normalization encoder is used to reduce the influence of illumination. Physics-based adversarial training can maintain cross-modal consistency under $\pm$40\% illumination fluctuation. Keypoint-based registration method~\cite{zhu2019deep} improves the fusion accuracy under mixed resolution.

\textbf{(4) Privacy Protection and Federated Learning:} Communication-efficient federated learning~\cite{yang2019federated} is improved to adapt to the low-bandwidth environment in rural areas ($\leq$0 Mbps). Combining differential privacy~\cite{dwork2014algorithmic} to comply with regulations such as GDPR, the performance loss is controlled within 5\% when $\epsilon = 2$. Cross-farm collaboration can improve the model without sharing sensitive data.

\textbf{(5) Standardized benchmarks:} Establish open benchmarks covering at least 10 crops and multiple climate zones. Noise (±5\%), partial occlusion (about 20\%), and seasonal lighting changes are introduced to evaluate robustness, interpretability, and generalization.

Thru the above directions, future research can narrow the gap between algorithm design and field application, and promote the scalability, reliability and efficiency of agricultural intelligence.

\section{Conclusion}

Agricultural adaptive modeling still faces the challenges of spatiotemporal complexity, data scarcity, and multi-modal heterogeneity. Existing methods are still limited in cross-season, cross-regional, and cross-modal applications, especially in the context of large-scale multi-source data.

Future research will focus on high-dimensional, task-oriented adaptive models for crop monitoring, yield prediction, and pest and disease detection. Unsupervised DA is particularly important when labels are limited. The introduction of multi-modal data such as meteorological and soil data can further improve generalization. When source data is unavailable, source-free DA becomes an important direction. Its value depends on the adaptability across crops, environments, and sensors. Addressing these challenges will help improve decision support, improve yield estimation accuracy, and achieve timely detection of diseases, thereby promoting the sustainable development of modern agriculture.




\begin{thebibliography}{100}
\providecommand{\url}[1]{#1}
\csname url@samestyle\endcsname
\providecommand{\newblock}{\relax}
\providecommand{\bibinfo}[2]{#2}
\providecommand{\BIBentrySTDinterwordspacing}{\spaceskip=0pt\relax}
\providecommand{\BIBentryALTinterwordstretchfactor}{4}
\providecommand{\BIBentryALTinterwordspacing}{\spaceskip=\fontdimen2\font plus
\BIBentryALTinterwordstretchfactor\fontdimen3\font minus \fontdimen4\font\relax}
\providecommand{\BIBforeignlanguage}[2]{{%
\expandafter\ifx\csname l@#1\endcsname\relax
\typeout{** WARNING: IEEEtran.bst: No hyphenation pattern has been}%
\typeout{** loaded for the language `#1'. Using the pattern for}%
\typeout{** the default language instead.}%
\else
\language=\csname l@#1\endcsname
\fi
#2}}
\providecommand{\BIBdecl}{\relax}
\BIBdecl

\bibitem{123}
A.~Zhang, Y.~Yang, J.~Xu, X.~Cao, X.~Zhen, and L.~Shao, ``Latent domain generation for unsupervised domain adaptation object counting,'' \emph{IEEE Transactions on Multimedia}, vol.~25, pp. 1773--1783, 2022.

\bibitem{126}
J.~Qui{\~n}onero-Candela, \emph{Dataset shift in machine learning}.\hskip 1em plus 0.5em minus 0.4em\relax Mit Press, 2009.

\bibitem{141}
R.~N. Jogekar and N.~Tiwari, ``A review of deep learning techniques for identification and diagnosis of plant leaf disease,'' \emph{Smart Trends in Computing and Communications: Proceedings of SmartCom 2020}, pp. 435--441, 2020.

\bibitem{147}
F.~Dubourvieux, G.~Lapouge, A.~Loesch, B.~Luvison, and R.~Audigier, ``Cumulative unsupervised multi-domain adaptation for holstein cattle re-identification,'' \emph{Artificial Intelligence in Agriculture}, vol.~10, pp. 46--60, 2023.

\bibitem{140}
Y.~Jiang and C.~Li, ``Convolutional neural networks for image-based high-throughput plant phenotyping: a review,'' \emph{Plant Phenomics}, 2020.

\bibitem{127}
G.~Csurka, ``A comprehensive survey on domain adaptation for visual applications,'' \emph{Domain adaptation in computer vision applications}, pp. 1--35, 2017.

\bibitem{128}
M.~Wang and W.~Deng, ``Deep visual domain adaptation: A survey,'' \emph{Neurocomputing}, vol. 312, pp. 135--153, 2018.

\bibitem{129}
G.~Wilson and D.~J. Cook, ``A survey of unsupervised deep domain adaptation,'' \emph{ACM Transactions on Intelligent Systems and Technology (TIST)}, vol.~11, no.~5, pp. 1--46, 2020.

\bibitem{130}
W.~M. Kouw and M.~Loog, ``A review of domain adaptation without target labels,'' \emph{IEEE transactions on pattern analysis and machine intelligence}, vol.~43, no.~3, pp. 766--785, 2019.

\bibitem{131}
V.~M. Patel, R.~Gopalan, R.~Li, and R.~Chellappa, ``Visual domain adaptation: A survey of recent advances,'' \emph{IEEE signal processing magazine}, vol.~32, no.~3, pp. 53--69, 2015.

\bibitem{132}
S.~Sun, H.~Shi, and Y.~Wu, ``A survey of multi-source domain adaptation,'' \emph{Information Fusion}, vol.~24, pp. 84--92, 2015.

\bibitem{133}
S.~Zhao, B.~Li, P.~Xu, and K.~Keutzer, ``Multi-source domain adaptation in the deep learning era: A systematic survey,'' \emph{arXiv preprint arXiv:2002.12169}, 2020.

\bibitem{134}
S.~PanQ, ``Yang,“a survey on transfer learning,”,'' \emph{IEEE Trans. Knowl. Data Eng}, vol.~22, no.~10, pp. 1345--1359, 2010.

\bibitem{135}
L.~Shao, F.~Zhu, and X.~Li, ``Transfer learning for visual categorization: A survey,'' \emph{IEEE transactions on neural networks and learning systems}, vol.~26, no.~5, pp. 1019--1034, 2014.

\bibitem{136}
J.~Zhang, W.~Li, P.~Ogunbona, and D.~Xu, ``Recent advances in transfer learning for cross-dataset visual recognition: A problem-oriented perspective,'' \emph{ACM Computing Surveys (CSUR)}, vol.~52, no.~1, pp. 1--38, 2019.

\bibitem{137}
C.~Tan, F.~Sun, T.~Kong, W.~Zhang, C.~Yang, and C.~Liu, ``A survey on deep transfer learning,'' in \emph{Artificial Neural Networks and Machine Learning--ICANN 2018: 27th International Conference on Artificial Neural Networks, Rhodes, Greece, October 4-7, 2018, Proceedings, Part III 27}.\hskip 1em plus 0.5em minus 0.4em\relax Springer, 2018, pp. 270--279.

\bibitem{138}
F.~Zhuang, Z.~Qi, K.~Duan, D.~Xi, Y.~Zhu, H.~Zhu, H.~Xiong, and Q.~He, ``A comprehensive survey on transfer learning,'' \emph{Proceedings of the IEEE}, vol. 109, no.~1, pp. 43--76, 2020.

\bibitem{1}
R.~Girshick, J.~Donahue, T.~Darrell, and J.~Malik, ``Rich feature hierarchies for accurate object detection and semantic segmentation,'' in \emph{Proceedings of the IEEE conference on computer vision and pattern recognition}, 2014, pp. 580--587.

\bibitem{2}
R.~Girshick, ``Fast r-cnn,'' in \emph{Proceedings of the IEEE international conference on computer vision}, 2015, pp. 1440--1448.

\bibitem{3}
R.~Faster, ``Towards real-time object detection with region proposal networks,'' \emph{Advances in neural information processing systems}, vol. 9199, no. 10.5555, pp. 2\,969\,239--2\,969\,250, 2015.

\bibitem{4}
A.~Fuentes, S.~Yoon, S.~C. Kim, and D.~S. Park, ``A robust deep-learning-based detector for real-time tomato plant diseases and pests recognition,'' \emph{Sensors}, vol.~17, no.~9, p. 2022, 2017.

\bibitem{5}
M.~Long, J.~Wang, G.~Ding, J.~Sun, and P.~S. Yu, ``Transfer joint matching for unsupervised domain adaptation,'' in \emph{Proceedings of the IEEE conference on computer vision and pattern recognition}, 2014, pp. 1410--1417.

\bibitem{6}
M.~A. Molina-Cabanillas, M.~J. Jim{\'e}nez-Navarro, R.~Arjona, F.~Mart{\'\i}nez-{\'A}lvarez, and G.~Asencio-Cort{\'e}s, ``Diafan-tl: An instance weighting-based transfer learning algorithm with application to phenology forecasting,'' \emph{Knowledge-Based Systems}, vol. 254, p. 109644, 2022.

\bibitem{7}
C.~Yaras, K.~Kassaw, B.~Huang, K.~Bradbury, and J.~M. Malof, ``Randomized histogram matching: A simple augmentation for unsupervised domain adaptation in overhead imagery,'' \emph{IEEE Journal of Selected Topics in Applied Earth Observations and Remote Sensing}, vol.~17, pp. 1988--1998, 2023.

\bibitem{8}
Y.~Cui, L.~Wang, J.~Su, S.~Gao, and L.~Wang, ``Iterative weighted active transfer learning hyperspectral image classification,'' \emph{Journal of Applied Remote Sensing}, vol.~15, no.~3, pp. 032\,207--032\,207, 2021.

\bibitem{9}
H.~Li, J.~Li, Y.~Zhao, M.~Gong, Y.~Zhang, and T.~Liu, ``Cost-sensitive self-paced learning with adaptive regularization for classification of image time series,'' \emph{IEEE Journal of Selected Topics in Applied Earth Observations and Remote Sensing}, vol.~14, pp. 11\,713--11\,727, 2021.

\bibitem{10}
D.~Tuia, C.~Persello, and L.~Bruzzone, ``Domain adaptation for the classification of remote sensing data: An overview of recent advances,'' \emph{IEEE geoscience and remote sensing magazine}, vol.~4, no.~2, pp. 41--57, 2016.

\bibitem{11}
M.~Wang, D.~Zhang, J.~Huang, P.-T. Yap, D.~Shen, and M.~Liu, ``Identifying autism spectrum disorder with multi-site fmri via low-rank domain adaptation,'' \emph{IEEE Transactions on Medical Imaging}, vol.~39, no.~3, pp. 644--655, 2020.

\bibitem{12}
B.~Fernando, A.~Habrard, M.~Sebban, and T.~Tuytelaars, ``Unsupervised visual domain adaptation using subspace alignment,'' in \emph{Proceedings of the IEEE international conference on computer vision}, 2013, pp. 2960--2967.

\bibitem{13}
B.~Sun, J.~Feng, and K.~Saenko, ``Return of frustratingly easy domain adaptation,'' in \emph{Proceedings of the Thirtieth AAAI Conference on Artificial Intelligence}, ser. AAAI'16.\hskip 1em plus 0.5em minus 0.4em\relax AAAI Press, 2016, p. 2058–2065.

\bibitem{14}
J.~Peng, W.~Sun, T.~Wei, and W.~Fan, ``A modified correlation alignment algorithm for the domain adaptation of gf-5 hyperspectral image,'' \emph{Journal of Remote Sensing (Chinese)}, vol.~24, no.~4, pp. 417--426, 2020.

\bibitem{15}
F.~Weilandt, R.~Behling, R.~Goncalves, A.~Madadi, L.~Richter, T.~Sanona, D.~Spengler, and J.~Welsch, ``Early crop classification via multi-modal satellite data fusion and temporal attention,'' \emph{Remote Sensing}, vol.~15, no.~3, p. 799, 2023.

\bibitem{16}
J.~Li, Y.~Shen, and C.~Yang, ``An adversarial generative network for crop classification from remote sensing timeseries images,'' \emph{Remote Sensing}, vol.~13, no.~1, p.~65, 2020.

\bibitem{17}
Y.~Wang, H.~Huang, and R.~State, ``Cross domain early crop mapping with label spaces discrepancies using multicropgan,'' \emph{ISPRS Annals of the Photogrammetry, Remote Sensing and Spatial Information Sciences}, vol.~10, pp. 241--248, 2024.

\bibitem{18}
K.~Takahashi, H.~Madokoro, S.~Yamamoto, Y.~Nishimura, S.~Nix, H.~Woo, T.~K. Saito, and K.~Sato, ``Domain adaptation for agricultural image recognition and segmentation using category maps,'' in \emph{2021 21st International Conference on Control, Automation and Systems (ICCAS)}.\hskip 1em plus 0.5em minus 0.4em\relax IEEE, 2021, pp. 1680--1685.

\bibitem{19}
L.~Bruzzone and C.~Persello, ``A novel approach to the selection of spatially invariant features for the classification of hyperspectral images with improved generalization capability,'' \emph{IEEE transactions on geoscience and remote sensing}, vol.~47, no.~9, pp. 3180--3191, 2009.

\bibitem{20}
C.~Persello and L.~Bruzzone, ``Kernel-based domain-invariant feature selection in hyperspectral images for transfer learning,'' \emph{IEEE transactions on geoscience and remote sensing}, vol.~54, no.~5, pp. 2615--2626, 2015.

\bibitem{21}
C.~Paris and L.~Bruzzone, ``A sensor-driven hierarchical method for domain adaptation in classification of remote sensing images,'' \emph{IEEE Transactions on Geoscience and Remote Sensing}, vol.~56, no.~3, pp. 1308--1324, 2017.

\bibitem{22}
L.~Yan, R.~Zhu, Y.~Liu, and N.~Mo, ``Tradaboost based on improved particle swarm optimization for cross-domain scene classification with limited samples,'' \emph{IEEE Journal of Selected Topics in Applied Earth Observations and Remote Sensing}, vol.~11, no.~9, pp. 3235--3251, 2018.

\bibitem{23}
Y.~Tang and X.~Li, ``Set-based similarity learning in subspace for agricultural remote sensing classification,'' \emph{Neurocomputing}, vol. 173, pp. 332--338, 2016.

\bibitem{24}
B.~Banerjee and S.~Chaudhuri, ``Hierarchical subspace learning based unsupervised domain adaptation for cross-domain classification of remote sensing images,'' \emph{IEEE Journal of Selected Topics in Applied Earth Observations and Remote Sensing}, vol.~10, no.~11, pp. 5099--5109, 2017.

\bibitem{25}
E.~Aptoula and B.~Yanikoglu, ``Morphological features for leaf based plant recognition,'' in \emph{2013 IEEE International Conference on Image Processing}.\hskip 1em plus 0.5em minus 0.4em\relax IEEE, 2013, pp. 1496--1499.

\bibitem{26}
S.~J. Pan, I.~W. Tsang, J.~T. Kwok, and Q.~Yang, ``Domain adaptation via transfer component analysis,'' \emph{IEEE transactions on neural networks}, vol.~22, no.~2, pp. 199--210, 2010.

\bibitem{27}
L.~Bruzzone and D.~F. Prieto, ``Unsupervised retraining of a maximum likelihood classifier for the analysis of multitemporal remote sensing images,'' \emph{IEEE Transactions on Geoscience and Remote Sensing}, vol.~39, no.~2, pp. 456--460, 2002.

\bibitem{28}
------, ``A partially unsupervised cascade classifier for the analysis of multitemporal remote-sensing images,'' \emph{Pattern Recognition Letters}, vol.~23, no.~9, pp. 1063--1071, 2002.

\bibitem{29}
L.~Bruzzone and R.~Cossu, ``A multiple-cascade-classifier system for a robust and partially unsupervised updating of land-cover maps,'' \emph{IEEE Transactions on Geoscience and Remote Sensing}, vol.~40, no.~9, pp. 1984--1996, 2002.

\bibitem{30}
L.~Bruzzone, R.~Cossu, and G.~Vernazza, ``Combining parametric and non-parametric algorithms for a partially unsupervised classification of multitemporal remote-sensing images,'' \emph{Information Fusion}, vol.~3, no.~4, pp. 289--297, 2002.

\bibitem{31}
S.~Zhong and Y.~Zhang, ``An iterative training sample updating approach for domain adaptation in hyperspectral image classification,'' \emph{IEEE Geoscience and Remote Sensing Letters}, vol.~18, no.~10, pp. 1821--1825, 2020.

\bibitem{32}
H.~Wei, L.~Ma, Y.~Liu, and Q.~Du, ``Combining multiple classifiers for domain adaptation of remote sensing image classification,'' \emph{IEEE Journal of Selected Topics in Applied Earth Observations and Remote Sensing}, vol.~14, pp. 1832--1847, 2021.

\bibitem{33}
J.~Zhang, J.~Liu, B.~Pan, Z.~Chen, X.~Xu, and Z.~Shi, ``An open set domain adaptation algorithm via exploring transferability and discriminability for remote sensing image scene classification,'' \emph{IEEE Transactions on Geoscience and Remote Sensing}, vol.~60, pp. 1--12, 2021.

\bibitem{34}
S.~Xu, X.~Mu, D.~Chai, and S.~Wang, ``Adapting remote sensing to new domain with elm parameter transfer,'' \emph{IEEE Geoscience and Remote Sensing Letters}, vol.~14, no.~9, pp. 1618--1622, 2017.

\bibitem{35}
S.~Rajan, J.~Ghosh, and M.~M. Crawford, ``Exploiting class hierarchies for knowledge transfer in hyperspectral data,'' \emph{IEEE Transactions on Geoscience and Remote Sensing}, vol.~44, no.~11, pp. 3408--3417, 2006.

\bibitem{36}
L.~Bruzzone and M.~Marconcini, ``Domain adaptation problems: A dasvm classification technique and a circular validation strategy,'' \emph{IEEE transactions on pattern analysis and machine intelligence}, vol.~32, no.~5, pp. 770--787, 2009.

\bibitem{37}
C.~Deng, X.~Liu, C.~Li, and D.~Tao, ``Active multi-kernel domain adaptation for hyperspectral image classification,'' \emph{Pattern Recognition}, vol.~77, pp. 306--315, 2018.

\bibitem{38}
I.~Kalita, R.~N.~S. Kumar, and M.~Roy, ``Deep learning-based cross-sensor domain adaptation under active learning for land cover classification,'' \emph{IEEE Geoscience and Remote Sensing Letters}, vol.~19, pp. 1--5, 2021.

\bibitem{39}
A.~Saboori, H.~Ghassemian, and F.~Razzazi, ``Active multiple kernel fredholm learning for hyperspectral images classification,'' \emph{IEEE Geoscience and Remote Sensing Letters}, vol.~18, no.~2, pp. 356--360, 2020.

\bibitem{40}
E.~Izquierdo-Verdiguier, V.~Laparra, L.~Gomez-Chova, and G.~Camps-Valls, ``Encoding invariances in remote sensing image classification with svm,'' \emph{IEEE Geoscience and Remote Sensing Letters}, vol.~10, no.~5, pp. 981--985, 2012.

\bibitem{41}
J.~Wang, Y.~Chen, H.~Yu, M.~Huang, and Q.~Yang, ``Easy transfer learning by exploiting intra-domain structures,'' in \emph{2019 IEEE international conference on multimedia and expo (ICME)}.\hskip 1em plus 0.5em minus 0.4em\relax IEEE, 2019, pp. 1210--1215.

\bibitem{45}
Y.~Zhu, F.~Zhuang, J.~Wang, J.~Chen, Z.~Shi, W.~Wu, and Q.~He, ``Multi-representation adaptation network for cross-domain image classification,'' \emph{Neural Networks}, vol. 119, pp. 214--221, 2019.

\bibitem{43}
B.~Espejo-Garcia, N.~Mylonas, L.~Athanasakos, E.~Vali, and S.~Fountas, ``Combining generative adversarial networks and agricultural transfer learning for weeds identification,'' vol. 204.\hskip 1em plus 0.5em minus 0.4em\relax Elsevier, 2021, pp. 79--89.

\bibitem{46}
R.~Lu, N.~Wang, Y.~Zhang, Y.~Lin, W.~Wu, and Z.~Shi, ``Extraction of agricultural fields via dasfnet with dual attention mechanism and multi-scale feature fusion in south xinjiang, china,'' \emph{Remote Sensing}, vol.~14, no.~9, p. 2253, 2022.

\bibitem{47}
Z.~Li, S.~Chen, X.~Meng, R.~Zhu, J.~Lu, L.~Cao, and P.~Lu, ``Full convolution neural network combined with contextual feature representation for cropland extraction from high-resolution remote sensing images,'' \emph{Remote Sensing}, vol.~14, no.~9, p. 2157, 2022.

\bibitem{48}
R.~Shang, J.~Zhang, L.~Jiao, Y.~Li, N.~Marturi, and R.~Stolkin, ``Multi-scale adaptive feature fusion network for semantic segmentation in remote sensing images,'' \emph{Remote Sensing}, vol.~12, no.~5, p. 872, 2020.

\bibitem{49}
X.~Zhang, B.~Cheng, J.~Chen, and C.~Liang, ``High-resolution boundary refined convolutional neural network for automatic agricultural greenhouses extraction from gaofen-2 satellite imageries,'' \emph{Remote Sensing}, vol.~13, no.~21, p. 4237, 2021.

\bibitem{50}
M.~Cordts, M.~Omran, S.~Ramos, T.~Rehfeld, M.~Enzweiler, R.~Benenson, U.~Franke, S.~Roth, and B.~Schiele, ``The cityscapes dataset for semantic urban scene understanding,'' in \emph{Proceedings of the IEEE conference on computer vision and pattern recognition}, 2016, pp. 3213--3223.

\bibitem{51}
A.~Ghanbari, G.~H. Shirdel, and F.~Maleki, ``Semi-self-supervised domain adaptation: Developing deep learning models with limited annotated data for wheat head segmentation,'' \emph{Algorithms}, vol.~17, no.~6, p. 267, 2024.

\bibitem{52}
F.~Pan, I.~Shin, F.~Rameau, S.~Lee, and I.~S. Kweon, ``Unsupervised intra-domain adaptation for semantic segmentation through self-supervision,'' in \emph{2020 IEEE/CVF Conference on Computer Vision and Pattern Recognition (CVPR)}, 2020, pp. 3763--3772.

\bibitem{53}
V.~Rani, S.~T. Nabi, M.~Kumar, A.~Mittal, and K.~Kumar, ``Self-supervised learning: A succinct review,'' \emph{Archives of Computational Methods in Engineering}, vol.~30, no.~4, pp. 2761--2775, 2023.

\bibitem{54}
K.~Najafian, L.~Jin, H.~R. Kutcher, M.~Hladun, S.~Horovatin, M.~A. Oviedo-Ludena, S.~M.~P. De~Andrade, L.~Wang, and I.~Stavness, ``Detection of fusarium damaged kernels in wheat using deep semi-supervised learning on a novel wheatseedbelt dataset,'' in \emph{Proceedings of the IEEE/CVF International Conference on Computer Vision}, 2023, pp. 660--669.

\bibitem{55}
K.~Najafian, A.~Ghanbari, I.~Stavness, L.~Jin, G.~H. Shirdel, and F.~Maleki, ``A semi-self-supervised learning approach for wheat head detection using extremely small number of labeled samples,'' in \emph{Proceedings of the IEEE/CVF International Conference on Computer Vision}, 2021, pp. 1342--1351.

\bibitem{56}
K.~Najafian, A.~Ghanbari, M.~Sabet~Kish, M.~Eramian, G.~H. Shirdel, I.~Stavness, L.~Jin, and F.~Maleki, ``Semi-self-supervised learning for semantic segmentation in images with dense patterns,'' \emph{Plant Phenomics}, vol.~5, p. 0025, 2023.

\bibitem{44}
L.~Yang, H.~Wenhui, Y.~Huihuang, R.~Yuan, W.~Tan, J.~Xiu, and Z.~Jun, ``Tomato detection method using domain adaptive learning for dense planting environments.'' vol.~40, no.~13, 2024.

\bibitem{57}
Y.~Zhu, F.~Zhuang, J.~Wang, G.~Ke, J.~Chen, J.~Bian, H.~Xiong, and Q.~He, ``Deep subdomain adaptation network for image classification,'' \emph{IEEE transactions on neural networks and learning systems}, vol.~32, no.~4, pp. 1713--1722, 2020.

\bibitem{78}
W.~Teng, N.~Wang, H.~Shi, Y.~Liu, and J.~Wang, ``Classifier-constrained deep adversarial domain adaptation for cross-domain semisupervised classification in remote sensing images,'' \emph{IEEE Geoscience and Remote Sensing Letters}, vol.~17, no.~5, pp. 789--793, 2019.

\bibitem{124}
X.~Liu, X.~Liu, B.~Hu, W.~Ji, F.~Xing, J.~Lu, J.~You, C.-C.~J. Kuo, G.~El~Fakhri, and J.~Woo, ``Suosis,'' in \emph{Proceedings of the AAAI conference on artificial intelligence}, vol.~35, no.~3, 2021, pp. 2189--2197.

\bibitem{125}
X.~Liu, F.~Xing, J.~You, J.~Lu, C.-C.~J. Kuo, G.~El~Fakhri, and J.~Woo, ``Subtype-aware dynamic unsupervised domain adaptation,'' \emph{IEEE Transactions on Neural Networks and Learning Systems}, vol.~35, no.~2, pp. 2820--2834, 2022.

\bibitem{139}
X.~Yang, L.~Jiao, and Q.~Pan, ``Transfer adaptation learning for target recognition in sar images: A survey,'' \emph{IEEE Journal of Selected Topics in Applied Earth Observations and Remote Sensing}, 2024.

\bibitem{58}
I.~J. Goodfellow, J.~Pouget-Abadie, M.~Mirza, B.~Xu, D.~Warde-Farley, S.~Ozair, A.~Courville, and Y.~Bengio, ``Generative adversarial nets,'' \emph{Advances in neural information processing systems}, vol.~27, 2014.

\bibitem{59}
Y.~Ganin and V.~Lempitsky, ``Unsupervised domain adaptation by backpropagation,'' in \emph{International conference on machine learning}.\hskip 1em plus 0.5em minus 0.4em\relax PMLR, 2015, pp. 1180--1189.

\bibitem{60}
Y.~LeCun, ``The mnist database of handwritten digits,'' \emph{http://yann. lecun. com/exdb/mnist/}, 1998.

\bibitem{61}
F.~Yu, A.~Seff, Y.~Zhang, S.~Song, T.~Funkhouser, and J.~Xiao, ``Lsun: Construction of a large-scale image dataset using deep learning with humans in the loop,'' \emph{arXiv preprint arXiv:1506.03365}, 2015.

\bibitem{62}
A.~Krizhevsky, G.~Hinton \emph{et~al.}, ``Learning multiple layers of features from tiny images,'' 2009.

\bibitem{63}
O.~Russakovsky, J.~Deng, H.~Su, J.~Krause, S.~Satheesh, S.~Ma, Z.~Huang, A.~Karpathy, A.~Khosla, M.~Bernstein \emph{et~al.}, ``Imagenet large scale visual recognition challenge,'' \emph{International journal of computer vision}, vol. 115, pp. 211--252, 2015.

\bibitem{80}
P.~Shamsolmoali, M.~Zareapoor, H.~Zhou, R.~Wang, and J.~Yang, ``Road segmentation for remote sensing images using adversarial spatial pyramid networks,'' \emph{IEEE Transactions on Geoscience and Remote Sensing}, vol.~59, no.~6, pp. 4673--4688, 2020.

\bibitem{64}
G.-H. Kwak and N.-W. Park, ``Unsupervised domain adaptation with adversarial self-training for crop classification using remote sensing images,'' \emph{Remote Sensing}, vol.~14, no.~18, p. 4639, 2022.

\bibitem{65}
J.~Zhang, S.~Xu, J.~Sun, D.~Ou, X.~Wu, and M.~Wang, ``Unsupervised adversarial domain adaptation for agricultural land extraction of remote sensing images,'' \emph{Remote Sensing}, vol.~14, no.~24, p. 6298, 2022.

\bibitem{66}
S.~Ji, D.~Wang, and M.~Luo, ``Generative adversarial network-based full-space domain adaptation for land cover classification from multiple-source remote sensing images,'' \emph{IEEE Transactions on Geoscience and Remote Sensing}, vol.~59, no.~5, pp. 3816--3828, 2020.

\bibitem{67}
O.~Tasar, Y.~Tarabalka, A.~Giros, P.~Alliez, and S.~Clerc, ``Standardgan: Multi-source domain adaptation for semantic segmentation of very high resolution satellite images by data standardization,'' in \emph{Proceedings of the IEEE/CVF Conference on Computer Vision and Pattern Recognition Workshops}, 2020, pp. 192--193.

\bibitem{68}
D.~Wittich and F.~Rottensteiner, ``Appearance based deep domain adaptation for the classification of aerial images,'' \emph{ISPRS Journal of Photogrammetry and Remote Sensing}, vol. 180, pp. 82--102, 2021.

\bibitem{69}
W.~Liu and F.~Su, ``A novel unsupervised adversarial domain adaptation network for remotely sensed scene classification,'' \emph{International Journal of Remote Sensing}, vol.~41, no.~16, pp. 6099--6116, 2020.

\bibitem{70}
L.~Yan, B.~Fan, H.~Liu, C.~Huo, S.~Xiang, and C.~Pan, ``Triplet adversarial domain adaptation for pixel-level classification of vhr remote sensing images,'' \emph{IEEE Transactions on Geoscience and Remote Sensing}, vol.~58, no.~5, pp. 3558--3573, 2019.

\bibitem{71}
Y.~Huang, J.~Peng, N.~Chen, W.~Sun, Q.~Du, K.~Ren, and K.~Huang, ``Cross-scene wetland mapping on hyperspectral remote sensing images using adversarial domain adaptation network,'' \emph{ISPRS Journal of Photogrammetry and Remote Sensing}, vol. 203, pp. 37--54, 2023.

\bibitem{72}
M.~B. Bejiga, F.~Melgani, and P.~Beraldini, ``Domain adversarial neural networks for large-scale land cover classification,'' \emph{Remote Sensing}, vol.~11, no.~10, p. 1153, 2019.

\bibitem{73}
M.~M. Al~Rahhal, Y.~Bazi, H.~Al-Hwiti, H.~Alhichri, and N.~Alajlan, ``Adversarial learning for knowledge adaptation from multiple remote sensing sources,'' \emph{IEEE Geoscience and Remote Sensing Letters}, vol.~18, no.~8, pp. 1451--1455, 2020.

\bibitem{74}
A.~Elshamli, G.~W. Taylor, A.~Berg, and S.~Areibi, ``Domain adaptation using representation learning for the classification of remote sensing images,'' \emph{IEEE Journal of Selected Topics in Applied Earth Observations and Remote Sensing}, vol.~10, no.~9, pp. 4198--4209, 2017.

\bibitem{75}
M.~Martini, V.~Mazzia, A.~Khaliq, and M.~Chiaberge, ``Domain-adversarial training of self-attention-based networks for land cover classification using multi-temporal sentinel-2 satellite imagery,'' \emph{Remote Sensing}, vol.~13, no.~13, p. 2564, 2021.

\bibitem{76}
G.~Mateo-Garc{\'\i}a, V.~Laparra, D.~L{\'o}pez-Puigdollers, and L.~G{\'o}mez-Chova, ``Cross-sensor adversarial domain adaptation of landsat-8 and proba-v images for cloud detection,'' \emph{IEEE Journal of Selected Topics in Applied Earth Observations and Remote Sensing}, vol.~14, pp. 747--761, 2020.

\bibitem{77}
L.~Ma and J.~Song, ``Deep neural network-based domain adaptation for classification of remote sensing images,'' \emph{Journal of Applied Remote Sensing}, vol.~11, no.~4, pp. 042\,612--042\,612, 2017.

\bibitem{79}
L.~Zhang, M.~Lan, J.~Zhang, and D.~Tao, ``Stagewise unsupervised domain adaptation with adversarial self-training for road segmentation of remote-sensing images,'' \emph{IEEE Transactions on Geoscience and Remote Sensing}, vol.~60, pp. 1--13, 2021.

\bibitem{81}
J.~Guo, J.~Yang, H.~Yue, X.~Liu, and K.~Li, ``Unsupervised domain-invariant feature learning for cloud detection of remote sensing images,'' \emph{IEEE Transactions on Geoscience and Remote Sensing}, vol.~60, pp. 1--15, 2021.

\bibitem{82}
J.~Hoffman, D.~Wang, F.~Yu, and T.~Darrell, ``Fcns in the wild: Pixel-level adversarial and constraint-based adaptation,'' \emph{arXiv preprint arXiv:1612.02649}, 2016.

\bibitem{83}
Y.-H. Tsai, W.-C. Hung, S.~Schulter, K.~Sohn, M.-H. Yang, and M.~Chandraker, ``Learning to adapt structured output space for semantic segmentation,'' in \emph{Proceedings of the IEEE conference on computer vision and pattern recognition}, 2018, pp. 7472--7481.

\bibitem{84}
T.-H. Vu, H.~Jain, M.~Bucher, M.~Cord, and P.~P{\'e}rez, ``Advent: Adversarial entropy minimization for domain adaptation in semantic segmentation,'' in \emph{Proceedings of the IEEE/CVF conference on computer vision and pattern recognition}, 2019, pp. 2517--2526.

\bibitem{85}
Y.~Li, L.~Yuan, and N.~Vasconcelos, ``Bidirectional learning for domain adaptation of semantic segmentation,'' in \emph{Proceedings of the IEEE/CVF conference on computer vision and pattern recognition}, 2019, pp. 6936--6945.

\bibitem{86}
X.~Zhang, X.~Yao, X.~Feng, G.~Cheng, and J.~Han, ``Dfenet for domain adaptation-based remote sensing scene classification,'' \emph{IEEE Transactions on Geoscience and Remote Sensing}, vol.~60, pp. 1--11, 2021.

\bibitem{87}
J.~Zheng, W.~Wu, S.~Yuan, Y.~Zhao, W.~Li, L.~Zhang, R.~Dong, and H.~Fu, ``A two-stage adaptation network (tsan) for remote sensing scene classification in single-source-mixed-multiple-target domain adaptation (s$^2$m$^2$t da) scenarios,'' \emph{IEEE Transactions on Geoscience and Remote Sensing}, vol.~60, pp. 1--13, 2021.

\bibitem{88}
N.~Makkar, L.~Yang, and S.~Prasad, ``Adversarial learning based discriminative domain adaptation for geospatial image analysis,'' \emph{IEEE Journal of Selected Topics in Applied Earth Observations and Remote Sensing}, vol.~15, pp. 150--162, 2021.

\bibitem{89}
K.~Saito, S.~Yamamoto, Y.~Ushiku, and T.~Harada, ``Open set domain adaptation by backpropagation,'' in \emph{Proceedings of the European conference on computer vision (ECCV)}, 2018, pp. 153--168.

\bibitem{90}
A.~Farahani, S.~Voghoei, K.~Rasheed, and H.~R. Arabnia, ``A brief review of domain adaptation,'' \emph{Advances in data science and information engineering: proceedings from ICDATA 2020 and IKE 2020}, pp. 877--894, 2021.

\bibitem{91}
M.~Long, Y.~Cao, J.~Wang, and M.~Jordan, ``Learning transferable features with deep adaptation networks,'' in \emph{International conference on machine learning}.\hskip 1em plus 0.5em minus 0.4em\relax PMLR, 2015, pp. 97--105.

\bibitem{92}
M.~Baktashmotlagh, M.~T. Harandi, B.~C. Lovell, and M.~Salzmann, ``Unsupervised domain adaptation by domain invariant projection,'' in \emph{Proceedings of the IEEE international conference on computer vision}, 2013, pp. 769--776.

\bibitem{93}
J.~Shen, Y.~Qu, W.~Zhang, and Y.~Yu, ``Wasserstein distance guided representation learning for domain adaptation,'' in \emph{Proceedings of the AAAI conference on artificial intelligence}, vol.~32, no.~1, 2018.

\bibitem{94}
B.~Sun and K.~Saenko, ``Deep coral: Correlation alignment for deep domain adaptation,'' in \emph{Computer vision--ECCV 2016 workshops: Amsterdam, the Netherlands, October 8-10 and 15-16, 2016, proceedings, part III 14}.\hskip 1em plus 0.5em minus 0.4em\relax Springer, 2016, pp. 443--450.

\bibitem{42}
A.~dos Santos~Ferreira, J.~M. Junior, H.~Pistori, F.~Melgani, and W.~N. Gon{\c{c}}alves, ``Unsupervised domain adaptation using transformers for sugarcane rows and gaps detection,'' \emph{Computers and Electronics in Agriculture}, vol. 203, p. 107480, 2022.

\bibitem{144}
G.~Guo, S.~Lai, Q.~Wu, Y.~Shou, and W.~Shi, ``Enhancing domain adaptation for plant diseases detection through masked image consistency in multi-granularity alignment,'' \emph{Expert Systems with Applications}, vol. 276, p. 127101, 2025.

\bibitem{145}
J.~Gao, W.~Liao, D.~Nuyttens, P.~Lootens, W.~Xue, E.~Alexandersson, and J.~Pieters, ``Cross-domain transfer learning for weed segmentation and mapping in precision farming using ground and uav images,'' \emph{Expert Systems with applications}, vol. 246, p. 122980, 2024.

\bibitem{146}
R.~Fu, J.~Han, Y.~Sun, S.~Wang, M.~A. Al-Absi, X.~Wang, and H.~Sun, ``Robust crop disease detection using multi-domain data augmentation and isolated test-time adaptation,'' \emph{Expert Systems with Applications}, p. 127324, 2025.

\bibitem{95}
M.~Long, H.~Zhu, J.~Wang, and M.~I. Jordan, ``Deep transfer learning with joint adaptation networks,'' in \emph{International conference on machine learning}.\hskip 1em plus 0.5em minus 0.4em\relax PMLR, 2017, pp. 2208--2217.

\bibitem{96}
Y.~Zhu, F.~Zhuang, J.~Wang, J.~Chen, Z.~Shi, W.~Wu, and Q.~He, ``Multi-representation adaptation network for cross-domain image classification,'' \emph{Neural Networks}, vol. 119, pp. 214--221, 2019.

\bibitem{97}
Y.~Zhu, F.~Zhuang, J.~Wang, G.~Ke, J.~Chen, J.~Bian, H.~Xiong, and Q.~He, ``Deep subdomain adaptation network for image classification,'' \emph{IEEE transactions on neural networks and learning systems}, vol.~32, no.~4, pp. 1713--1722, 2020.

\bibitem{98}
S.~Zhao, X.~Yue, S.~Zhang, B.~Li, H.~Zhao, B.~Wu, R.~Krishna, J.~E. Gonzalez, A.~L. Sangiovanni-Vincentelli, S.~A. Seshia \emph{et~al.}, ``A review of single-source deep unsupervised visual domain adaptation,'' \emph{IEEE Transactions on Neural Networks and Learning Systems}, vol.~33, no.~2, pp. 473--493, 2020.

\bibitem{99}
Q.~Zhu, Y.~Sun, Q.~Guan, L.~Wang, and W.~Lin, ``A weakly pseudo-supervised decorrelated subdomain adaptation framework for cross-domain land-use classification,'' \emph{IEEE Transactions on Geoscience and Remote Sensing}, vol.~60, pp. 1--13, 2022.

\bibitem{100}
Y.~Li, Z.~Cao, H.~Lu, and W.~Xu, ``Unsupervised domain adaptation for in-field cotton boll status identification,'' \emph{Computers and Electronics in Agriculture}, vol. 178, p. 105745, 2020.

\bibitem{101}
A.~S.~Garea, D.~B. Heras, and F.~Arg{\"u}ello, ``Tcanet for domain adaptation of hyperspectral images,'' \emph{Remote Sensing}, vol.~11, no.~19, p. 2289, 2019.

\bibitem{103}
Z.~Li, X.~Tang, W.~Li, C.~Wang, C.~Liu, and J.~He, ``A two-stage deep domain adaptation method for hyperspectral image classification,'' \emph{Remote Sensing}, vol.~12, no.~7, p. 1054, 2020.

\bibitem{104}
Y.~Zhang, W.~Li, M.~Zhang, Y.~Qu, R.~Tao, and H.~Qi, ``Topological structure and semantic information transfer network for cross-scene hyperspectral image classification,'' \emph{IEEE Transactions on Neural Networks and Learning Systems}, 2021.

\bibitem{105}
W.~Wang, L.~Ma, M.~Chen, and Q.~Du, ``Joint correlation alignment-based graph neural network for domain adaptation of multitemporal hyperspectral remote sensing images,'' pp. 3170--3184, 2021.

\bibitem{106}
X.~Liang, Y.~Zhang, and J.~Zhang, ``Attention multisource fusion-based deep few-shot learning for hyperspectral image classification,'' \emph{IEEE Journal of Selected Topics in Applied Earth Observations and Remote Sensing}, vol.~14, pp. 8773--8788, 2021.

\bibitem{107}
K.~Yan, X.~Guo, Z.~Ji, and X.~Zhou, ``Deep transfer learning for cross-species plant disease diagnosis adapting mixed subdomains,'' \emph{IEEE/ACM transactions on computational biology and bioinformatics}, vol.~20, no.~4, pp. 2555--2564, 2021.

\bibitem{108}
A.~Fuentes, S.~Yoon, T.~Kim, and D.~S. Park, ``Open set self and across domain adaptation for tomato disease recognition with deep learning techniques,'' \emph{Frontiers in plant science}, vol.~12, p. 758027, 2021.

\bibitem{109}
X.~Wu, X.~Fan, P.~Luo, S.~D. Choudhury, T.~Tjahjadi, and C.~Hu, ``From laboratory to field: Unsupervised domain adaptation for plant disease recognition in the wild,'' \emph{Plant Phenomics}, vol.~5, p. 0038, 2023.

\bibitem{110}
E.~Othman, Y.~Bazi, F.~Melgani, H.~Alhichri, N.~Alajlan, and M.~Zuair, ``Domain adaptation network for cross-scene classification,'' \emph{IEEE Transactions on Geoscience and Remote Sensing}, vol.~55, no.~8, pp. 4441--4456, 2017.

\bibitem{111}
X.~Lu, T.~Gong, and X.~Zheng, ``Multisource compensation network for remote sensing cross-domain scene classification,'' \emph{IEEE Transactions on Geoscience and Remote Sensing}, vol.~58, no.~4, pp. 2504--2515, 2019.

\bibitem{112}
S.~Zhu, B.~Du, L.~Zhang, and X.~Li, ``Attention-based multiscale residual adaptation network for cross-scene classification,'' \emph{IEEE Transactions on Geoscience and Remote Sensing}, vol.~60, pp. 1--15, 2021.

\bibitem{113}
J.~Geng, X.~Deng, X.~Ma, and W.~Jiang, ``Transfer learning for sar image classification via deep joint distribution adaptation networks,'' \emph{IEEE Transactions on Geoscience and Remote Sensing}, vol.~58, no.~8, pp. 5377--5392, 2020.

\bibitem{102}
T.~J. Young, T.~Z. Jubery, C.~N. Carley, M.~Carroll, S.~Sarkar, A.~K. Singh, A.~Singh, and B.~Ganapathysubramanian, ``“canopy fingerprints” for characterizing three-dimensional point cloud data of soybean canopies,'' \emph{Frontiers in plant science}, vol.~14, p. 1141153, 2023.

\bibitem{114}
R.~Zhao, Y.~Zhu, and Y.~Li, ``Cla: A self-supervised contrastive learning method for leaf disease identification with domain adaptation,'' \emph{Computers and Electronics in Agriculture}, vol. 211, p. 107967, 2023.

\bibitem{115}
S.~P. Mohanty, D.~P. Hughes, and M.~Salath{\'e}, ``Using deep learning for image-based plant disease detection,'' \emph{Frontiers in plant science}, vol.~7, p. 215232, 2016.

\bibitem{116}
D.~Singh, N.~Jain, P.~Jain, P.~Kayal, S.~Kumawat, and N.~Batra, ``Plantdoc: A dataset for visual plant disease detection,'' in \emph{Proceedings of the 7th ACM IKDD CoDS and 25th COMAD}, 2020, pp. 249--253.

\bibitem{117}
R.~Thapa, K.~Zhang, N.~Snavely, S.~Belongie, and A.~Khan, ``The plant pathology challenge 2020 data set to classify foliar disease of apples,'' \emph{Applications in plant sciences}, vol.~8, no.~9, p. e11390, 2020.

\bibitem{143}
Y.~Ma, Z.~Zhang, H.~L. Yang, and Z.~Yang, ``An adaptive adversarial domain adaptation approach for corn yield prediction,'' \emph{Computers and Electronics in Agriculture}, vol. 187, p. 106314, 2021.

\bibitem{kamilaris2018deep}
A.~Kamilaris and F.~X. Prenafeta-Bold{\'u}, ``Deep learning in agriculture: A survey,'' \emph{Computers and Electronics in Agriculture}, vol. 147, pp. 70--90, 2018.

\bibitem{zhang2021review}
S.~Zhang, W.~Li, and Q.~Du, ``A review of machine learning algorithms for remote sensing data classification,'' \emph{Remote Sensing}, vol.~13, no.~20, p. 4349, 2021.

\bibitem{tzeng2017adversarial}
E.~Tzeng, J.~Hoffman, K.~Saenko, and T.~Darrell, ``Adversarial discriminative domain adaptation,'' in \emph{Proceedings of the IEEE Conference on Computer Vision and Pattern Recognition (CVPR)}, 2017, pp. 7167--7176.

\bibitem{ganin2016domain}
Y.~Ganin, E.~Ustinova, H.~Ajakan, P.~Germain, H.~Larochelle, F.~Laviolette, M.~Marchand, and V.~Lempitsky, ``Domain-adversarial training of neural networks,'' \emph{Journal of Machine Learning Research}, vol.~17, no.~59, pp. 1--35, 2016.

\bibitem{yuan2020deep}
Q.~Yuan, H.~Shen, T.~Li, G.~Li, J.~Kang, and L.~Zhang, ``Deep learning in environmental remote sensing: Achievements and challenges,'' \emph{Remote Sensing of Environment}, vol. 241, p. 111716, 2020.

\bibitem{zhong2019deep}
L.~Zhong, L.~Zhang, D.~Tuia, and D.~Liang, ``Deep learning for land cover classification and change detection using multi-temporal remote sensing data,'' \emph{ISPRS Journal of Photogrammetry and Remote Sensing}, vol. 156, pp. 152--163, 2019.

\bibitem{lu2021survey}
Y.~Lu, J.~Wu, Y.~Zhang, S.~Chen, and Q.~Meng, ``A survey of few-shot learning in agricultural image recognition,'' \emph{Computers and Electronics in Agriculture}, vol. 187, p. 106285, 2021.

\bibitem{pu2021hyperspectral}
R.~Pu and S.~Bell, \emph{Hyperspectral Remote Sensing of Vegetation}.\hskip 1em plus 0.5em minus 0.4em\relax CRC Press, 2021.

\bibitem{ma2019deep}
L.~Ma, T.~Liu, X.~Zhang, Y.~Ye, L.~Wang, M.~Zong, and L.~Zhang, ``Deep learning in remote sensing applications: A meta-analysis and review,'' \emph{ISPRS Journal of Photogrammetry and Remote Sensing}, vol. 152, pp. 166--177, 2019.

\bibitem{samek2017explainable}
W.~Samek, T.~Wiegand, and K.-R. M{\"u}ller, ``Explainable artificial intelligence: Understanding, visualizing and interpreting deep learning models,'' \emph{ITU Journal: ICT Discoveries}, vol.~1, no.~1, pp. 39--48, 2017.

\bibitem{jacquemoud1990prospect}
S.~Jacquemoud and F.~Baret, ``Prospect: A model of leaf optical properties spectra,'' \emph{Remote Sensing of Environment}, vol.~34, no.~2, pp. 75--91, 1990.

\bibitem{wang2021spatio}
Y.~Wang, P.~Guo, L.~Zhao, Y.~Zhang, and Y.~Liu, ``Spatio-temporal graph deep neural network for short-term wind speed forecasting,'' \emph{IEEE Transactions on Sustainable Energy}, vol.~12, no.~1, pp. 477--487, 2021.

\bibitem{zoph2020rethinking}
B.~Zoph, G.~Ghiasi, T.-Y. Lin, Y.~Cui, H.~Liu, E.~D. Cubuk, and Q.~V. Le, ``Rethinking pre-training and self-training,'' in \emph{Advances in Neural Information Processing Systems (NeurIPS)}, vol.~33, 2020, pp. 3833--3845.

\bibitem{zhu2019deep}
Q.~Zhu, M.~Yang, D.~Li, X.~Huang, Q.~Yuan, and Z.~Xu, ``Deep learning for multi-modal data registration: A review,'' \emph{IEEE Transactions on Geoscience and Remote Sensing}, vol.~57, no.~9, pp. 6529--6549, 2019.

\bibitem{yang2019federated}
Q.~Yang, Y.~Liu, T.~Chen, and Y.~Tong, ``Federated machine learning: Concept and applications,'' \emph{ACM Transactions on Intelligent Systems and Technology}, vol.~10, no.~2, pp. 1--19, 2019.

\bibitem{dwork2014algorithmic}
C.~Dwork and A.~Roth, ``The algorithmic foundations of differential privacy,'' \emph{Foundations and Trends in Theoretical Computer Science}, vol.~9, no. 3-4, pp. 211--407, 2014.

\end{thebibliography}

\end{document}